\documentclass[acmlarge, nonacm]{acmart}

\def\BibTeX{{\rm B\kern-.05em{\sc i\kern-.025em b}\kern-.08emT\kern-.1667em\lower.7ex\hbox{E}\kern-.125emX}}
    
\usepackage{todonotes}
\usepackage{multirow}
\usepackage{natbib}
\usepackage{subcaption}
\usepackage{tablefootnote}

\usepackage{enumitem}

\makeatletter
\def\mdseries@tt{m}
\makeatother
\usepackage[draft]{minted}

\makeatletter
\def\PYG@reset{\let\PYG@it=\relax \let\PYG@bf=\relax%
    \let\PYG@ul=\relax \let\PYG@tc=\relax%
    \let\PYG@bc=\relax \let\PYG@ff=\relax}
\def\PYG@tok#1{\csname PYG@tok@#1\endcsname}
\def\PYG@toks#1+{\ifx\relax#1\empty\else%
    \PYG@tok{#1}\expandafter\PYG@toks\fi}
\def\PYG@do#1{\PYG@bc{\PYG@tc{\PYG@ul{%
    \PYG@it{\PYG@bf{\PYG@ff{#1}}}}}}}
\def\PYG#1#2{\PYG@reset\PYG@toks#1+\relax+\PYG@do{#2}}

\expandafter\def\csname PYG@tok@gd\endcsname{\def\PYG@tc##1{\textcolor[rgb]{0.63,0.00,0.00}{##1}}}
\expandafter\def\csname PYG@tok@gu\endcsname{\let\PYG@bf=\textbf\def\PYG@tc##1{\textcolor[rgb]{0.50,0.00,0.50}{##1}}}
\expandafter\def\csname PYG@tok@gt\endcsname{\def\PYG@tc##1{\textcolor[rgb]{0.00,0.27,0.87}{##1}}}
\expandafter\def\csname PYG@tok@gs\endcsname{\let\PYG@bf=\textbf}
\expandafter\def\csname PYG@tok@gr\endcsname{\def\PYG@tc##1{\textcolor[rgb]{1.00,0.00,0.00}{##1}}}
\expandafter\def\csname PYG@tok@cm\endcsname{\let\PYG@it=\textit\def\PYG@tc##1{\textcolor[rgb]{0.25,0.50,0.50}{##1}}}
\expandafter\def\csname PYG@tok@vg\endcsname{\def\PYG@tc##1{\textcolor[rgb]{0.10,0.09,0.49}{##1}}}
\expandafter\def\csname PYG@tok@m\endcsname{\def\PYG@tc##1{\textcolor[rgb]{0.40,0.40,0.40}{##1}}}
\expandafter\def\csname PYG@tok@mh\endcsname{\def\PYG@tc##1{\textcolor[rgb]{0.40,0.40,0.40}{##1}}}
\expandafter\def\csname PYG@tok@go\endcsname{\def\PYG@tc##1{\textcolor[rgb]{0.53,0.53,0.53}{##1}}}
\expandafter\def\csname PYG@tok@ge\endcsname{\let\PYG@it=\textit}
\expandafter\def\csname PYG@tok@vc\endcsname{\def\PYG@tc##1{\textcolor[rgb]{0.10,0.09,0.49}{##1}}}
\expandafter\def\csname PYG@tok@il\endcsname{\def\PYG@tc##1{\textcolor[rgb]{0.40,0.40,0.40}{##1}}}
\expandafter\def\csname PYG@tok@cs\endcsname{\let\PYG@it=\textit\def\PYG@tc##1{\textcolor[rgb]{0.25,0.50,0.50}{##1}}}
\expandafter\def\csname PYG@tok@cp\endcsname{\def\PYG@tc##1{\textcolor[rgb]{0.74,0.48,0.00}{##1}}}
\expandafter\def\csname PYG@tok@gi\endcsname{\def\PYG@tc##1{\textcolor[rgb]{0.00,0.63,0.00}{##1}}}
\expandafter\def\csname PYG@tok@gh\endcsname{\let\PYG@bf=\textbf\def\PYG@tc##1{\textcolor[rgb]{0.00,0.00,0.50}{##1}}}
\expandafter\def\csname PYG@tok@ni\endcsname{\let\PYG@bf=\textbf\def\PYG@tc##1{\textcolor[rgb]{0.60,0.60,0.60}{##1}}}
\expandafter\def\csname PYG@tok@nl\endcsname{\def\PYG@tc##1{\textcolor[rgb]{0.63,0.63,0.00}{##1}}}
\expandafter\def\csname PYG@tok@nn\endcsname{\let\PYG@bf=\textbf\def\PYG@tc##1{\textcolor[rgb]{0.00,0.00,1.00}{##1}}}
\expandafter\def\csname PYG@tok@no\endcsname{\def\PYG@tc##1{\textcolor[rgb]{0.53,0.00,0.00}{##1}}}
\expandafter\def\csname PYG@tok@na\endcsname{\def\PYG@tc##1{\textcolor[rgb]{0.49,0.56,0.16}{##1}}}
\expandafter\def\csname PYG@tok@nb\endcsname{\def\PYG@tc##1{\textcolor[rgb]{0.00,0.50,0.00}{##1}}}
\expandafter\def\csname PYG@tok@nc\endcsname{\let\PYG@bf=\textbf\def\PYG@tc##1{\textcolor[rgb]{0.00,0.00,1.00}{##1}}}
\expandafter\def\csname PYG@tok@nd\endcsname{\def\PYG@tc##1{\textcolor[rgb]{0.67,0.13,1.00}{##1}}}
\expandafter\def\csname PYG@tok@ne\endcsname{\let\PYG@bf=\textbf\def\PYG@tc##1{\textcolor[rgb]{0.82,0.25,0.23}{##1}}}
\expandafter\def\csname PYG@tok@nf\endcsname{\def\PYG@tc##1{\textcolor[rgb]{0.00,0.00,1.00}{##1}}}
\expandafter\def\csname PYG@tok@si\endcsname{\let\PYG@bf=\textbf\def\PYG@tc##1{\textcolor[rgb]{0.73,0.40,0.53}{##1}}}
\expandafter\def\csname PYG@tok@s2\endcsname{\def\PYG@tc##1{\textcolor[rgb]{0.73,0.13,0.13}{##1}}}
\expandafter\def\csname PYG@tok@vi\endcsname{\def\PYG@tc##1{\textcolor[rgb]{0.10,0.09,0.49}{##1}}}
\expandafter\def\csname PYG@tok@nt\endcsname{\let\PYG@bf=\textbf\def\PYG@tc##1{\textcolor[rgb]{0.00,0.50,0.00}{##1}}}
\expandafter\def\csname PYG@tok@nv\endcsname{\def\PYG@tc##1{\textcolor[rgb]{0.10,0.09,0.49}{##1}}}
\expandafter\def\csname PYG@tok@s1\endcsname{\def\PYG@tc##1{\textcolor[rgb]{0.73,0.13,0.13}{##1}}}
\expandafter\def\csname PYG@tok@kd\endcsname{\let\PYG@bf=\textbf\def\PYG@tc##1{\textcolor[rgb]{0.00,0.50,0.00}{##1}}}
\expandafter\def\csname PYG@tok@sh\endcsname{\def\PYG@tc##1{\textcolor[rgb]{0.73,0.13,0.13}{##1}}}
\expandafter\def\csname PYG@tok@sc\endcsname{\def\PYG@tc##1{\textcolor[rgb]{0.73,0.13,0.13}{##1}}}
\expandafter\def\csname PYG@tok@sx\endcsname{\def\PYG@tc##1{\textcolor[rgb]{0.00,0.50,0.00}{##1}}}
\expandafter\def\csname PYG@tok@bp\endcsname{\def\PYG@tc##1{\textcolor[rgb]{0.00,0.50,0.00}{##1}}}
\expandafter\def\csname PYG@tok@c1\endcsname{\let\PYG@it=\textit\def\PYG@tc##1{\textcolor[rgb]{0.25,0.50,0.50}{##1}}}
\expandafter\def\csname PYG@tok@kc\endcsname{\let\PYG@bf=\textbf\def\PYG@tc##1{\textcolor[rgb]{0.00,0.50,0.00}{##1}}}
\expandafter\def\csname PYG@tok@c\endcsname{\let\PYG@it=\textit\def\PYG@tc##1{\textcolor[rgb]{0.25,0.50,0.50}{##1}}}
\expandafter\def\csname PYG@tok@mf\endcsname{\def\PYG@tc##1{\textcolor[rgb]{0.40,0.40,0.40}{##1}}}
\expandafter\def\csname PYG@tok@err\endcsname{\def\PYG@bc##1{\setlength{\fboxsep}{0pt}\fcolorbox[rgb]{1.00,0.00,0.00}{1,1,1}{\strut ##1}}}
\expandafter\def\csname PYG@tok@mb\endcsname{\def\PYG@tc##1{\textcolor[rgb]{0.40,0.40,0.40}{##1}}}
\expandafter\def\csname PYG@tok@ss\endcsname{\def\PYG@tc##1{\textcolor[rgb]{0.10,0.09,0.49}{##1}}}
\expandafter\def\csname PYG@tok@sr\endcsname{\def\PYG@tc##1{\textcolor[rgb]{0.73,0.40,0.53}{##1}}}
\expandafter\def\csname PYG@tok@mo\endcsname{\def\PYG@tc##1{\textcolor[rgb]{0.40,0.40,0.40}{##1}}}
\expandafter\def\csname PYG@tok@kn\endcsname{\let\PYG@bf=\textbf\def\PYG@tc##1{\textcolor[rgb]{0.00,0.50,0.00}{##1}}}
\expandafter\def\csname PYG@tok@mi\endcsname{\def\PYG@tc##1{\textcolor[rgb]{0.40,0.40,0.40}{##1}}}
\expandafter\def\csname PYG@tok@gp\endcsname{\let\PYG@bf=\textbf\def\PYG@tc##1{\textcolor[rgb]{0.00,0.00,0.50}{##1}}}
\expandafter\def\csname PYG@tok@o\endcsname{\def\PYG@tc##1{\textcolor[rgb]{0.40,0.40,0.40}{##1}}}
\expandafter\def\csname PYG@tok@kr\endcsname{\let\PYG@bf=\textbf\def\PYG@tc##1{\textcolor[rgb]{0.00,0.50,0.00}{##1}}}
\expandafter\def\csname PYG@tok@s\endcsname{\def\PYG@tc##1{\textcolor[rgb]{0.73,0.13,0.13}{##1}}}
\expandafter\def\csname PYG@tok@kp\endcsname{\def\PYG@tc##1{\textcolor[rgb]{0.00,0.50,0.00}{##1}}}
\expandafter\def\csname PYG@tok@w\endcsname{\def\PYG@tc##1{\textcolor[rgb]{0.73,0.73,0.73}{##1}}}
\expandafter\def\csname PYG@tok@kt\endcsname{\def\PYG@tc##1{\textcolor[rgb]{0.69,0.00,0.25}{##1}}}
\expandafter\def\csname PYG@tok@ow\endcsname{\let\PYG@bf=\textbf\def\PYG@tc##1{\textcolor[rgb]{0.67,0.13,1.00}{##1}}}
\expandafter\def\csname PYG@tok@sb\endcsname{\def\PYG@tc##1{\textcolor[rgb]{0.73,0.13,0.13}{##1}}}
\expandafter\def\csname PYG@tok@k\endcsname{\let\PYG@bf=\textbf\def\PYG@tc##1{\textcolor[rgb]{0.00,0.50,0.00}{##1}}}
\expandafter\def\csname PYG@tok@se\endcsname{\let\PYG@bf=\textbf\def\PYG@tc##1{\textcolor[rgb]{0.73,0.40,0.13}{##1}}}
\expandafter\def\csname PYG@tok@sd\endcsname{\let\PYG@it=\textit\def\PYG@tc##1{\textcolor[rgb]{0.73,0.13,0.13}{##1}}}

\makeatother

\makeatletter
\def\PYGdefault@reset{\let\PYGdefault@it=\relax \let\PYGdefault@bf=\relax%
    \let\PYGdefault@ul=\relax \let\PYGdefault@tc=\relax%
    \let\PYGdefault@bc=\relax \let\PYGdefault@ff=\relax}
\def\PYGdefault@tok#1{\csname PYGdefault@tok@#1\endcsname}
\def\PYGdefault@toks#1+{\ifx\relax#1\empty\else%
    \PYGdefault@tok{#1}\expandafter\PYGdefault@toks\fi}
\def\PYGdefault@do#1{\PYGdefault@bc{\PYGdefault@tc{\PYGdefault@ul{%
    \PYGdefault@it{\PYGdefault@bf{\PYGdefault@ff{#1}}}}}}}
\def\PYGdefault#1#2{\PYGdefault@reset\PYGdefault@toks#1+\relax+\PYGdefault@do{#2}}

\expandafter\def\csname PYGdefault@tok@gd\endcsname{\def\PYGdefault@tc##1{\textcolor[rgb]{0.63,0.00,0.00}{##1}}}
\expandafter\def\csname PYGdefault@tok@gu\endcsname{\let\PYGdefault@bf=\textbf\def\PYGdefault@tc##1{\textcolor[rgb]{0.50,0.00,0.50}{##1}}}
\expandafter\def\csname PYGdefault@tok@gt\endcsname{\def\PYGdefault@tc##1{\textcolor[rgb]{0.00,0.27,0.87}{##1}}}
\expandafter\def\csname PYGdefault@tok@gs\endcsname{\let\PYGdefault@bf=\textbf}
\expandafter\def\csname PYGdefault@tok@gr\endcsname{\def\PYGdefault@tc##1{\textcolor[rgb]{1.00,0.00,0.00}{##1}}}
\expandafter\def\csname PYGdefault@tok@cm\endcsname{\let\PYGdefault@it=\textit\def\PYGdefault@tc##1{\textcolor[rgb]{0.25,0.50,0.50}{##1}}}
\expandafter\def\csname PYGdefault@tok@vg\endcsname{\def\PYGdefault@tc##1{\textcolor[rgb]{0.10,0.09,0.49}{##1}}}
\expandafter\def\csname PYGdefault@tok@m\endcsname{\def\PYGdefault@tc##1{\textcolor[rgb]{0.40,0.40,0.40}{##1}}}
\expandafter\def\csname PYGdefault@tok@mh\endcsname{\def\PYGdefault@tc##1{\textcolor[rgb]{0.40,0.40,0.40}{##1}}}
\expandafter\def\csname PYGdefault@tok@go\endcsname{\def\PYGdefault@tc##1{\textcolor[rgb]{0.53,0.53,0.53}{##1}}}
\expandafter\def\csname PYGdefault@tok@ge\endcsname{\let\PYGdefault@it=\textit}
\expandafter\def\csname PYGdefault@tok@vc\endcsname{\def\PYGdefault@tc##1{\textcolor[rgb]{0.10,0.09,0.49}{##1}}}
\expandafter\def\csname PYGdefault@tok@il\endcsname{\def\PYGdefault@tc##1{\textcolor[rgb]{0.40,0.40,0.40}{##1}}}
\expandafter\def\csname PYGdefault@tok@cs\endcsname{\let\PYGdefault@it=\textit\def\PYGdefault@tc##1{\textcolor[rgb]{0.25,0.50,0.50}{##1}}}
\expandafter\def\csname PYGdefault@tok@cp\endcsname{\def\PYGdefault@tc##1{\textcolor[rgb]{0.74,0.48,0.00}{##1}}}
\expandafter\def\csname PYGdefault@tok@gi\endcsname{\def\PYGdefault@tc##1{\textcolor[rgb]{0.00,0.63,0.00}{##1}}}
\expandafter\def\csname PYGdefault@tok@gh\endcsname{\let\PYGdefault@bf=\textbf\def\PYGdefault@tc##1{\textcolor[rgb]{0.00,0.00,0.50}{##1}}}
\expandafter\def\csname PYGdefault@tok@ni\endcsname{\let\PYGdefault@bf=\textbf\def\PYGdefault@tc##1{\textcolor[rgb]{0.60,0.60,0.60}{##1}}}
\expandafter\def\csname PYGdefault@tok@nl\endcsname{\def\PYGdefault@tc##1{\textcolor[rgb]{0.63,0.63,0.00}{##1}}}
\expandafter\def\csname PYGdefault@tok@nn\endcsname{\let\PYGdefault@bf=\textbf\def\PYGdefault@tc##1{\textcolor[rgb]{0.00,0.00,1.00}{##1}}}
\expandafter\def\csname PYGdefault@tok@no\endcsname{\def\PYGdefault@tc##1{\textcolor[rgb]{0.53,0.00,0.00}{##1}}}
\expandafter\def\csname PYGdefault@tok@na\endcsname{\def\PYGdefault@tc##1{\textcolor[rgb]{0.49,0.56,0.16}{##1}}}
\expandafter\def\csname PYGdefault@tok@nb\endcsname{\def\PYGdefault@tc##1{\textcolor[rgb]{0.00,0.50,0.00}{##1}}}
\expandafter\def\csname PYGdefault@tok@nc\endcsname{\let\PYGdefault@bf=\textbf\def\PYGdefault@tc##1{\textcolor[rgb]{0.00,0.00,1.00}{##1}}}
\expandafter\def\csname PYGdefault@tok@nd\endcsname{\def\PYGdefault@tc##1{\textcolor[rgb]{0.67,0.13,1.00}{##1}}}
\expandafter\def\csname PYGdefault@tok@ne\endcsname{\let\PYGdefault@bf=\textbf\def\PYGdefault@tc##1{\textcolor[rgb]{0.82,0.25,0.23}{##1}}}
\expandafter\def\csname PYGdefault@tok@nf\endcsname{\def\PYGdefault@tc##1{\textcolor[rgb]{0.00,0.00,1.00}{##1}}}
\expandafter\def\csname PYGdefault@tok@si\endcsname{\let\PYGdefault@bf=\textbf\def\PYGdefault@tc##1{\textcolor[rgb]{0.73,0.40,0.53}{##1}}}
\expandafter\def\csname PYGdefault@tok@s2\endcsname{\def\PYGdefault@tc##1{\textcolor[rgb]{0.73,0.13,0.13}{##1}}}
\expandafter\def\csname PYGdefault@tok@vi\endcsname{\def\PYGdefault@tc##1{\textcolor[rgb]{0.10,0.09,0.49}{##1}}}
\expandafter\def\csname PYGdefault@tok@nt\endcsname{\let\PYGdefault@bf=\textbf\def\PYGdefault@tc##1{\textcolor[rgb]{0.00,0.50,0.00}{##1}}}
\expandafter\def\csname PYGdefault@tok@nv\endcsname{\def\PYGdefault@tc##1{\textcolor[rgb]{0.10,0.09,0.49}{##1}}}
\expandafter\def\csname PYGdefault@tok@s1\endcsname{\def\PYGdefault@tc##1{\textcolor[rgb]{0.73,0.13,0.13}{##1}}}
\expandafter\def\csname PYGdefault@tok@kd\endcsname{\let\PYGdefault@bf=\textbf\def\PYGdefault@tc##1{\textcolor[rgb]{0.00,0.50,0.00}{##1}}}
\expandafter\def\csname PYGdefault@tok@sh\endcsname{\def\PYGdefault@tc##1{\textcolor[rgb]{0.73,0.13,0.13}{##1}}}
\expandafter\def\csname PYGdefault@tok@sc\endcsname{\def\PYGdefault@tc##1{\textcolor[rgb]{0.73,0.13,0.13}{##1}}}
\expandafter\def\csname PYGdefault@tok@sx\endcsname{\def\PYGdefault@tc##1{\textcolor[rgb]{0.00,0.50,0.00}{##1}}}
\expandafter\def\csname PYGdefault@tok@bp\endcsname{\def\PYGdefault@tc##1{\textcolor[rgb]{0.00,0.50,0.00}{##1}}}
\expandafter\def\csname PYGdefault@tok@c1\endcsname{\let\PYGdefault@it=\textit\def\PYGdefault@tc##1{\textcolor[rgb]{0.25,0.50,0.50}{##1}}}
\expandafter\def\csname PYGdefault@tok@kc\endcsname{\let\PYGdefault@bf=\textbf\def\PYGdefault@tc##1{\textcolor[rgb]{0.00,0.50,0.00}{##1}}}
\expandafter\def\csname PYGdefault@tok@c\endcsname{\let\PYGdefault@it=\textit\def\PYGdefault@tc##1{\textcolor[rgb]{0.25,0.50,0.50}{##1}}}
\expandafter\def\csname PYGdefault@tok@mf\endcsname{\def\PYGdefault@tc##1{\textcolor[rgb]{0.40,0.40,0.40}{##1}}}
\expandafter\def\csname PYGdefault@tok@err\endcsname{\def\PYGdefault@bc##1{\setlength{\fboxsep}{0pt}\fcolorbox[rgb]{1.00,0.00,0.00}{1,1,1}{\strut ##1}}}
\expandafter\def\csname PYGdefault@tok@mb\endcsname{\def\PYGdefault@tc##1{\textcolor[rgb]{0.40,0.40,0.40}{##1}}}
\expandafter\def\csname PYGdefault@tok@ss\endcsname{\def\PYGdefault@tc##1{\textcolor[rgb]{0.10,0.09,0.49}{##1}}}
\expandafter\def\csname PYGdefault@tok@sr\endcsname{\def\PYGdefault@tc##1{\textcolor[rgb]{0.73,0.40,0.53}{##1}}}
\expandafter\def\csname PYGdefault@tok@mo\endcsname{\def\PYGdefault@tc##1{\textcolor[rgb]{0.40,0.40,0.40}{##1}}}
\expandafter\def\csname PYGdefault@tok@kn\endcsname{\let\PYGdefault@bf=\textbf\def\PYGdefault@tc##1{\textcolor[rgb]{0.00,0.50,0.00}{##1}}}
\expandafter\def\csname PYGdefault@tok@mi\endcsname{\def\PYGdefault@tc##1{\textcolor[rgb]{0.40,0.40,0.40}{##1}}}
\expandafter\def\csname PYGdefault@tok@gp\endcsname{\let\PYGdefault@bf=\textbf\def\PYGdefault@tc##1{\textcolor[rgb]{0.00,0.00,0.50}{##1}}}
\expandafter\def\csname PYGdefault@tok@o\endcsname{\def\PYGdefault@tc##1{\textcolor[rgb]{0.40,0.40,0.40}{##1}}}
\expandafter\def\csname PYGdefault@tok@kr\endcsname{\let\PYGdefault@bf=\textbf\def\PYGdefault@tc##1{\textcolor[rgb]{0.00,0.50,0.00}{##1}}}
\expandafter\def\csname PYGdefault@tok@s\endcsname{\def\PYGdefault@tc##1{\textcolor[rgb]{0.73,0.13,0.13}{##1}}}
\expandafter\def\csname PYGdefault@tok@kp\endcsname{\def\PYGdefault@tc##1{\textcolor[rgb]{0.00,0.50,0.00}{##1}}}
\expandafter\def\csname PYGdefault@tok@w\endcsname{\def\PYGdefault@tc##1{\textcolor[rgb]{0.73,0.73,0.73}{##1}}}
\expandafter\def\csname PYGdefault@tok@kt\endcsname{\def\PYGdefault@tc##1{\textcolor[rgb]{0.69,0.00,0.25}{##1}}}
\expandafter\def\csname PYGdefault@tok@ow\endcsname{\let\PYGdefault@bf=\textbf\def\PYGdefault@tc##1{\textcolor[rgb]{0.67,0.13,1.00}{##1}}}
\expandafter\def\csname PYGdefault@tok@sb\endcsname{\def\PYGdefault@tc##1{\textcolor[rgb]{0.73,0.13,0.13}{##1}}}
\expandafter\def\csname PYGdefault@tok@k\endcsname{\let\PYGdefault@bf=\textbf\def\PYGdefault@tc##1{\textcolor[rgb]{0.00,0.50,0.00}{##1}}}
\expandafter\def\csname PYGdefault@tok@se\endcsname{\let\PYGdefault@bf=\textbf\def\PYGdefault@tc##1{\textcolor[rgb]{0.73,0.40,0.13}{##1}}}
\expandafter\def\csname PYGdefault@tok@sd\endcsname{\let\PYGdefault@it=\textit\def\PYGdefault@tc##1{\textcolor[rgb]{0.73,0.13,0.13}{##1}}}

\makeatother

\begin{document}
\title{Random Search and Reproducibility for Neural Architecture Search}

\author{Liam Li}
\email{me@liamcli.com}
\affiliation{%
  \institution{Carnegie Mellon University}
}

\author{Ameet Talwalkar}
\email{talwalkar@cmu.edu}
\affiliation{%
  \institution{Carnegie Mellon University and Determined AI}
}

\renewcommand{\shortauthors}{Li and Talwalkar}

\begin{abstract}
Neural architecture search (NAS) 
is a promising research direction that 
has the potential to 
replace expert-designed networks with learned, task-specific architectures.  
In this work, in order to help ground the empirical results in this field,  we  propose new NAS baselines that build off the following observations: (i) NAS is 
a specialized hyperparameter optimization problem; and (ii) random search is a competitive baseline for hyperparameter optimization. Leveraging these observations, 
we evaluate both random search with early-stopping and a novel random search with weight-sharing algorithm on two standard NAS benchmarks---PTB and CIFAR-10.  Our results show that random search with early-stopping is a competitive NAS baseline, e.g., it performs at least as well as ENAS~\citep{pham18ENAS}, a leading NAS method, on both benchmarks.
Additionally, random search with weight-sharing outperforms random search with early-stopping, achieving a state-of-the-art NAS result on PTB and a highly competitive result on CIFAR-10.
Finally, we explore the existing reproducibility issues of published NAS results. We note the lack of source material needed to exactly reproduce these results, and further discuss the robustness of published results given the various 
sources of variability in NAS experimental setups. Relatedly, we provide all information (code, random seeds, documentation) needed to exactly reproduce our results, and report our random search with weight-sharing results for each benchmark on multiple runs.
\end{abstract}


\maketitle

\section{Introduction}
\label{sec:intro}
Deep learning offers the promise of bypassing the process of manual feature engineering by learning representations in conjunction with statistical models in an end-to-end fashion. However, neural network architectures themselves are typically designed by experts in a painstaking, ad-hoc fashion. Neural architecture search (NAS) presents a promising path for alleviating this pain by automatically identifying architectures that are superior to hand-designed ones. Since the 
work by \citet{nasRL}, there has been explosion of research activity on this problem~\citep{pnas2018, liu2018hierarchical, Negrinho2017,Elsken2018pareto,Real2018,bender2018understanding,Jin2018,brock2018smash,pham18ENAS,zhang2018graph,liu2018darts,xie2018snas,cai2018proxylessnas}. 
Notably, there has been great industry interest in NAS, as evidenced by the vast computational~\citep{nasRL,Zoph2018LearningTA,Real2018} and marketing resources~\citep{automl} committed to industry-driven NAS research.
However, despite a steady stream of promising empirical results \citep{nasRL,Zoph2018LearningTA, Real2018, liu2018darts, Luo2018, cai2018proxylessnas},  
we see three fundamental issues with the current state of NAS research:

    \textbf{Inadequate Baselines.} Leading NAS methods exploit many of the strategies that were initially explored in the context of traditional hyperparameter optimization tasks, e.g., evolutionary search \citep{olson2016tpot,jaderberg2017pbt}, Bayesian optimization \citep{snoek2012practical,Bergstra2011, Hutter2011}, and gradient-based approaches \citep{Bengio2000,maclaurin2015gradient}. 
    Moreover, the NAS problem is in fact a specialized instance  of the broader hyperparameter optimization problem.  However, in spite of the  close relationship between these two problems, existing comparisons between novel NAS methods and standard hyperparameter optimization methods are inadequate.  In particular, to the best of our knowledge, no state-of-the-art hyperparameter optimization methods have been evaluated on standard NAS benchmarks.  \emph{Without benchmarking against leading hyperparameter optimization baselines, it difficult to quantify the performance gains provided by specialized NAS methods.}
   
    \textbf{Complex Methods.} 
    We have witnessed a proliferation of novel NAS methods, with research
    progressing in many different directions.  New approaches  introduce a significant amount of
    algorithmic complexity in the search process, including complicated training routines \citep{bender2018understanding, pham18ENAS,xie2018snas,cai2018proxylessnas}, architecture transformations \citep{wei2016morphism,Real2018,cai2018path, liu2018hierarchical, Elsken2018pareto}, and modeling assumptions \citep{Jin2018, Kandasamy2018, zhang2018graph, brock2018smash, pnas2018}  (see Figure~\ref{fig:components} and Section~\ref{ssec:background} for more details).
    While many technically diverse NAS methods demonstrate good empirical performance, they often lack corresponding ablation studies \citep{Luo2018,zhang2018graph,cai2018proxylessnas}, and as a result,  
    \emph{it is unclear what NAS component(s)
    are necessary to achieve a competitive empirical result}. 
    
   \textbf{Lack of Reproducibility.} 
   Experimental reproducibility is of paramount importance in the context of NAS research, given the empirical nature of the field, the complexity of new NAS methods, and the steep computational costs associated with empirical evaluation.  In particular, there are (at least) two important notions of reproducibility to consider: (1) ``exact'' reproducibility i.e., whether it is possible to reproduce explicitly reported experimental results; and ``broad'' reproducibility, i.e., the degree to which the reported experimental results are themselves robust and generalizable.  Broad reproducibility is difficult to measure due to the 
   computational burden of NAS methods and the high variance associated with extremal statistics.  However, most of the published results in this field do not even satisfy exact reproducibility.  \emph{For example, of the 12 papers published since 2018 at NeurIPS, ICML, and ICLR that introduce novel NAS methods (see Table~\ref{tab:reproduce}), none are exactly reproducible.} Indeed, each fails on account of some combination of missing model evaluation code, architecture search code, random seeds used for search and evaluation, and/or undocumented hyperparameter tuning.\footnote{It is important to note that these works vary drastically in terms of what materials they provide, and some authors such as 
   \citet{liu2018darts},  provide a relatively complete codebase for their methods.  However, even in the case of DARTS, the code for the CIFAR-10 benchmark is not deterministic and \citet{liu2018darts} do not provide random seeds or documentation regarding the post-processing steps in which they perform hyperparameter optimization on final architectures returned by DARTS. We were thus not able to reproduce the results in \citet{liu2018darts}, but we were able to use the DARTS code repository (\url{https://github.com/quark0/darts}) as the launching point for our experimental setup. }
 
While addressing these challenges will require community-wide efforts,
 in this work we present results that aim to make some initial progress on each of these issues. In particular, our contributions are as follows:
\begin{enumerate}[leftmargin=*]
    \item We help ground existing NAS results by providing a new perspective on the gap between traditional hyperparameter optimization and leading NAS methods. Specifically, we evaluate a general hyperparameter optimization method combining random search with early-stopping \citep{asha} on two standard NAS benchmarks (CIFAR-10 and PTB).  With approximately the same amount of compute as DARTS \citep{liu2018darts}, a state-of-the-art (SOTA) NAS method, this simple method provides a much more competitive baseline for both benchmarks: (1) on PTB, random search with early-stopping reaches test perplexity of 56.4 compared to the published result for ENAS \citep{pham18ENAS}, a leading NAS method, of 56.3,\footnote{We could not reproduce this result using the initial final architecture and code provided by the authors (\url{https://github.com/melodyguan/enas}).  They have since released another repository  (\url{https://github.com/google-research/google-research/tree/master/enas_lm}) that reports reproduced results but we have not verified these figures.} and (2) for CIFAR-10, random search with early-stopping achieves a test error of 2.85\%, whereas the published result for ENAS is 2.89\%.  While SOTA NAS methods like DARTS still outperform this baseline, our results demonstrate that the gap is not nearly as large as that suggested by published random search baselines on these tasks \citep{pham18ENAS,liu2018darts}.
    \item  We identify a small subset of NAS components that are sufficient for achieving good empirical results.  We construct a simple algorithm from the ground up starting from vanilla random search, 
    and demonstrate that properly tuned random search with weight-sharing is competitive with much more complicated methods when using similar computational budgets.  In particular, we identify the following meta-hyperparameters that impact the behavior of our algorithm: batch size, number of epochs, network size, and number of evaluated architectures. We  evaluate our proposed method using the same search space and evaluation scheme as DARTS \citep{liu2018darts}, a leading NAS method. We explore a few modifications of the meta-hyperparameters to improve search quality and make full use of available GPU memory and computational resources, and observe SOTA performance on the PTB benchmark and comparable performance to DARTS on the CIFAR-10 benchmark.  We emphasize that we do not perform additional hyperparameter tuning of the final architectures discovered at the end of the search process.
    \item 
    We open-source all of the necessary code, random seeds, and documentation necessary to reproduce our experiments.
    Our single machine results shown in Table~\ref{tab:rnn_sota} and Table~\ref{tab:cnn_sota} follow a deterministic experimental setup, given a fixed random seed, and satisfy exact reproducibility.  For these experiments on the two standard benchmarks, we study the broad reproduciblity of our random search with weight-sharing results by repeating our experiments with different random seeds.
    We observe non-trivial differences across independent runs and identify potential sources for these differences. Our results highlight the need for more careful reporting of experimental results, increased transparency of intermediate results, and more robust statistics to quantify the performance of NAS methods. 
    
\end{enumerate}

\subsection{Background}
\label{ssec:background}
\begin{figure}
    \centering
    \includegraphics[width=0.8\textwidth, trim=100 80 100 250, clip]{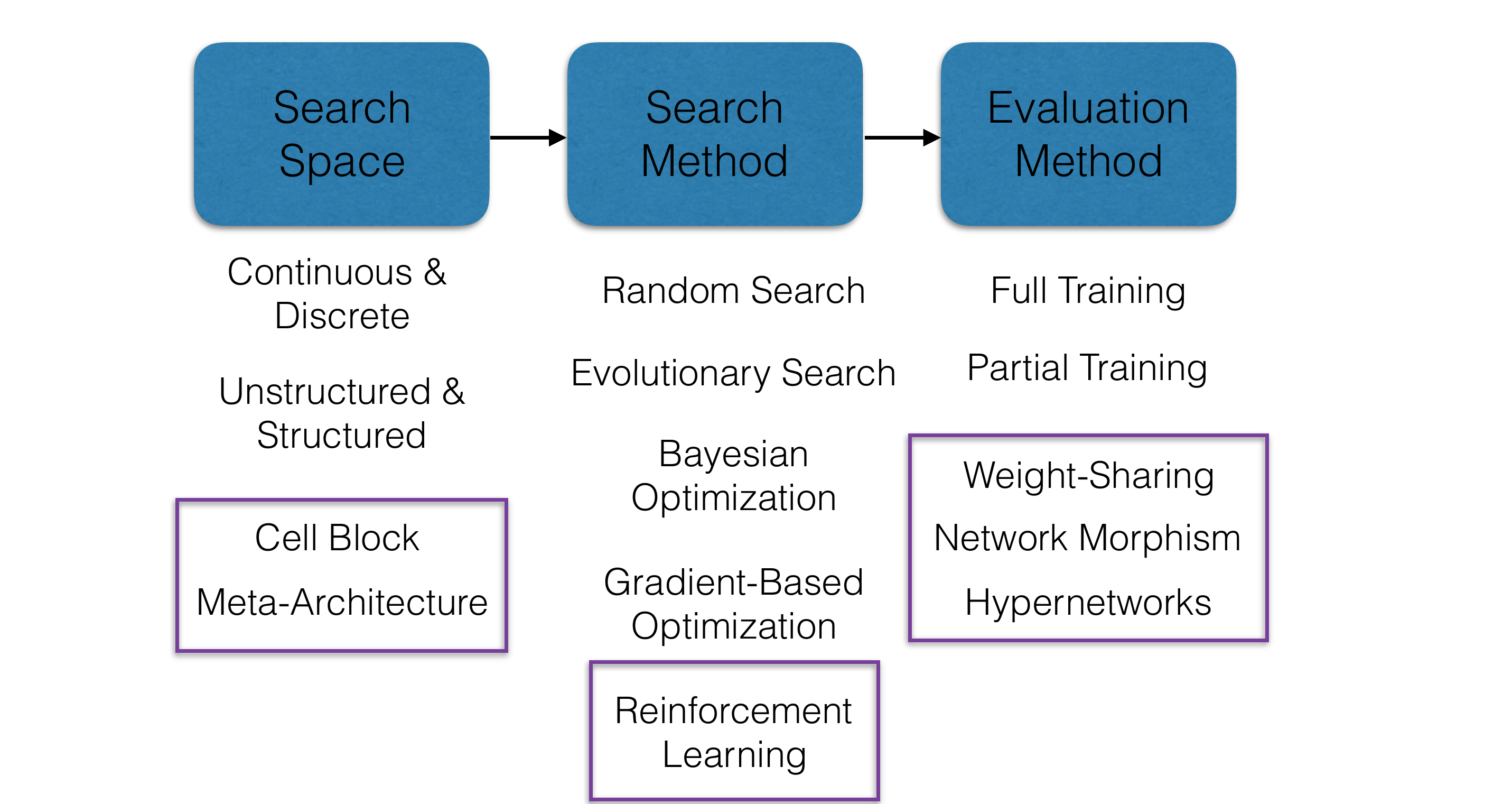}
    \caption{\textbf{Components of hyperparameter optimization.}  Primarily NAS-specific methods are outlined in purple.}
    \label{fig:components}
\end{figure}

We first provide an overview of the components of hyperparameter optimization and, by association, NAS.  
As shown in Figure~\ref{fig:components}, a general hyperparameter optimization problem has three components, each of which can have NAS-specific approaches.  We provide a brief overview of the components below, drawing attention to NAS-specific methods (see the survey by \citet{Elsken2018survey} for a more thorough coverage of NAS).  

\textbf{Search Space.} Hyperparameter optimization involves identifying a good hyperparameter configuration from a set of possible configurations. The search space defines this set of configurations,
and can include continuous or discrete hyperparameters in a structured or unstructured fashion~\citep{snoek2012practical,bergstra2012random,feurer2015efficient,olson2016tpot}.
NAS-specific search spaces usually involve discrete hyperparameters with additional structure that can be captured with a directed acyclic graph (DAG) \citep{pham18ENAS, liu2018darts}.    Additionally, since a search space for designing an entire architecture would have too many nodes and edges, search spaces are usually defined over some smaller building block, i.e., cell blocks, that are repeated in some way via a preset or learned meta-architecture to form a larger architecture \citep{Elsken2018survey}.  We design our random search NAS algorithm for such a cell block search space, using the same search spaces for the CIFAR-10 and PTB benchmarks as DARTS for our experiments in Section~\ref{sec:exp}. See Section~\ref{sec:alg} for a concrete example of one such search space. 

\textbf{Search Method.} Given a search space, there are various search methods to select putative configurations to evaluate. Random search is the most basic approach, yet it is quite effective in practice~\cite{bergstra2012random,hyperband}. Various general and NAS-specific adaptive methods 
have also been introduced, all of which      
attempt to bias the search in some way towards configurations that are more likely to perform well.  
In traditional hyperparameter optimization, the choice of search method can depends on the search space.  Bayesian approaches based on Gaussian processes \citep{snoek2012practical,fabolas2016, Swersky2013Multi,kandasamy2016gaussian} and gradient-based approaches~\citep{Bengio2000,maclaurin2015gradient} are generally only applicable to continuous search spaces.  In contrast, tree-based Bayesian \citep{Hutter2011, Bergstra2011}, evolutionary strategies \citep{olson2016tpot}, and random search are more flexible and can be applied to any search space. 
NAS-specific search methods can also be categorized into the same broad categories but are tailored for structured NAS search spaces (see Section~\ref{ssec:complex} for a more involved discussion). 

\textbf{Evaluation Method.} For each hyperparameter configuration considered by a search method, we must evaluate its quality. The default approach to perform such an evaluation involves fully training a model with the given hyperparameters, and subsequently measuring its quality, e.g., its predictive accuracy on a validation set. The first generation of NAS methods relied on full training evaluation, and thus required thousands of GPU days to achieve a desired result~\citep{nasRL,eshp2017,Zoph2018LearningTA, Real2018}.
In contrast, partial training methods exploit early-stopping to speed up the evaluation process at the cost of noisy estimates of configuration quality. These methods use Bayesian optimization \citep{fabolas2016, kandasamy2016gaussian,falkner2018bohb}, performance prediction \citep{vizier2017,earlystopping2015}, or multi-armed bandits \citep{JamiesonTalwalkar2015,hyperband,asha} to adaptively allocate resources to different configurations.
NAS-specific evaluation methods
exploit the structure of neural networks to provide even cheaper, heuristic estimates of quality. 
Many of these methods center around sharing and reuse: network morphisms build upon previously trained architectures \citep{cai2018path,Elsken2018pareto,Jin2018}; hypernetworks and performance prediction encode information from previously seen architectures \citep{brock2018smash,pnas2018,zhang2018graph}; and weight-sharing methods \citep{pham18ENAS, liu2018darts, bender2018understanding,xie2018snas,cai2018proxylessnas} use a single set of weights for all possible architectures.  

\section{Related Work}
\label{sec:related}
We now provide additional context for the three issues we identified with the current state of NAS research in Section~\ref{sec:intro}.  

\subsection{Inadequate Baselines}
Existing works in NAS do not provide adequate comparison to random search and other hyperparameter optimization methods.  Some works either compare to random search given a budget of just of few evaluations \citep{pham18ENAS,liu2018darts} or Bayesian optimization methods without efficient architecture evaluation schemes \citep{Jin2018}.  While \citet{Real2018} and \citet{cai2018path} provide a thorough comparison to random search, they use random search with full training even though partial training methods have been shown to be orders-of-magnitude faster than standard random search \citep{hyperband, asha}.  

While certain hyperparameter optimization methods \citep{snoek2012practical, maclaurin2015gradient, fabolas2016} require non-trivial modification in order to work with NAS search spaces, others are easily applicable to NAS problems \citep{Hutter2011, Bergstra2011, earlystopping2015,falkner2018bohb,hyperband,asha}.  Of these applicable methods, we choose to use a simple method combining random search with early-stopping called ASHA \citep{asha} to provide a competitive baseline for standard hyperparameter optimization. \citet{asha} showed ASHA to be a state-of-the-art, theoretically principled, bandit-based partial training method that outperforms leading adaptive search strategies for hyperparameter optimization.  We compare the empirical performance of ASHA with that of NAS methods in Section~\ref{sec:exp}.

\subsection{Complex Methods}
\label{ssec:complex}
Much of the complexity of NAS methods is introduced in the process of adapting search methods for NAS-specific search spaces: evolutionary approaches need to define a set of possible mutations to apply to different architectures \citep{eshp2017,Real2018}; Bayesian optimization approaches \citep{Jin2018,Kandasamy2018} rely on specially designed kernels; gradient-based methods transform the discrete architecture search problem into a continuous optimization problem so that gradients can be applied \citep{Luo2018, liu2018darts, xie2018snas,cai2018proxylessnas}; and \citet{nasRL}, \citet{Zoph2018LearningTA}, and \citet{pham18ENAS} use reinforcement learning to train a recurrent neural network controller to generate good architectures.  All of these search approaches add a significant amount of complexity with no clear winner, especially since methods some times use different search spaces and evaluation methods.  To simplify the search process and help isolate important components of NAS, we use random search to sample architectures from the search space.

Additional complexity is also introduced by the NAS-specific evaluation methods mentioned previously.
Network morphisms require architecture transformations that satisfy certain criteria; hypernetworks and performance prediction methods encode information from previously seen architectures in an auxiliary network; and weight-sharing methods \citep{pham18ENAS, liu2018darts, bender2018understanding,xie2018snas,cai2018proxylessnas} use a single set of weights for all possible architectures and hence, can require careful training routines.  
Despite their complexity, these more efficient NAS evaluation methods are 1-3 orders-of-magnitude cheaper than full training (see Table~\ref{tab:cnn_sota} and Table~\ref{tab:rnn_sota}),  at the expense of decreased fidelity to the true performance.  
Of these evaluation methods, network morphism still requires on the order of 100 GPU days \citep{pnas2018,Elsken2018pareto} and, while hypernetworks and prediction performance based methods can be cheaper, weight-sharing is less complex since it does not require training an auxiliary network.   In addition to the computational efficiency of weight-sharing methods \citep{liu2018darts, pham18ENAS, cai2018proxylessnas, xie2018snas}, which only require computation on the order of fully training a single architecture, this approach  has also achieved the best result on the two standard benchmarks \citep{liu2018darts,cai2018proxylessnas}.  Hence, we use random search with weight-sharing as our starting point for a simple and efficient NAS method.

Our work is inspired by the result of \citet{bender2018understanding}, which showed that random search, combined with a well-trained set of shared weights can successfully differentiate good architectures from poor performing ones. However, their work required several modifications to stabilize training (e.g., a tunable path dropout schedule over edges of the search DAG and a specialized ghost batch normalization scheme \citep{Hoffer2017TrainLG}).   Furthermore, they only report experimental results on the CIFAR-10 benchmark, on which they fell slightly short of the results for leading NAS methods. In contrast, our combination of random search with weight-sharing greatly simplifies the training routine and we identify key variables needed to achieve competitive results on both CIFAR-10 and PTB benchmarks.

\subsection{Lack of Reproducibility}
The earliest NAS results lacked exact and broad reproducibility due to the tremendous amount of computation required to achieve the results \citep{nasRL,Zoph2018LearningTA, Real2018}.  Additionally, some of these methods used specialized hardware (i.e., TPUs) that were not easily accessible 
to researchers at the time \citep{Real2018}.  Although the final architectures were eventually provided \citep{nasnetcode, amoebanetcode}, the code for the search methods used to produce these results has not been released, precluding researchers from reproducing these results even if they had sufficient computational resources. 

\begin{table}[h]
    \centering
    \small
    \caption{\textbf{Reproducibility of NAS Publications.}  Summary of the reproducibility status of recent NAS publications appearing in top machine learning conferences.  
    For the hyperparameter tuning column, N/A indicates we are not aware that the authors performed additional hyperparameter optimization.  
    \\\hspace{\textwidth}
    $^\dagger$ Published result is not reproducible for the PTB benchmark when training the reported final architecture with provided code.
    \\\hspace{\textwidth}
    $^*$ Code to reproduce experiments was requested on OpenReview.
    }
    \label{tab:reproduce}
    \begin{tabular}{cccccc}
    \hline
          &  & \textbf{Architecture} & \textbf{Model Evaluation } & \textbf{Random } & \textbf{Hyperparameter} \\
     \textbf{Conference} & \textbf{Publication} & \textbf{Search Code} & \textbf{Code} &\textbf{ Seeds} & \textbf{Tuning} \\
         \hline 
         ICLR 2018 &  \citet{brock2018smash} & Yes & Yes & No & N/A \\
         & \citet{liu2018hierarchical} & No & No & & \\
         \hline
         ICML 2018 & \citet{pham18ENAS}$^\dagger$ & Yes & Yes & No & Undocumented  \\
         & \citet{cai2018path} & Yes & Yes & No & N/A \\
         & \citet{bender2018understanding} & No & No & & \\
         \hline
         NIPS 2018 & \citet{Kandasamy2018} & Yes & Yes & No & N/A \\
         & \citet{Luo2018} & Yes & Yes & No & Grid Search \\
         \hline
         ICLR 2019 & \citet{liu2018darts} & Yes & Yes & No & Undocumented \\
         & \citet{cai2018proxylessnas} & No & Yes & No & N/A \\
         &  \citet{zhang2018graph}$^*$ & No & No & &  \\
         &  \citet{xie2018snas}$^*$ & No & No & &  \\
         &  \citet{cao2018learnable} & No & No & & \\
    \end{tabular}
    
\end{table}

Recently, it has become feasible to evaluate the exact and broad reproducibility of many SOTA methods due to their reduced computational cost.  However, while many authors have released code for their work \citep[e.g.,][]{pham18ENAS, liu2018darts, brock2018smash, cai2018path}, others have not made their code publicly available \citep[e.g.,][]{xie2018snas, zhang2018graph}, including the work most closely related to ours by \citet{bender2018understanding}.  We summarize the reproducibility of recent NAS publications at some of the major machine learning conferences in Table~\ref{tab:reproduce} according to the availability of the following:
\begin{enumerate}
    \item \textbf{Architecture search code.}  The output of this code is the final architecture that should be trained on the evaluation task.  
    \item \textbf{Model evaluation code.} The output of this code is the final performance on the evaluation task.  
    \item \textbf{Hyperparameter tuning documentation.}  This includes code used to perform hyperparameter tuning of the final architectures, if any.  
    \item \textbf{Random Seeds.} This includes random seeds used for both the search and post-processing (i.e., retraining of final architecture as well as any additional hyperparameter tuning) phases.  Most works provide the final architectures but random seeds are required to verify that the search process actually results in those final architectures and the performance of the final architectures matches the published result.  Note the random seeds are only useful if the code for search and post-processing phases are deterministic up to a random seed; this was not the case for the DARTS code used for the CIFAR-10 benchmark.
\end{enumerate}

All 4 criteria are necessary for exact reproducibility.  Due to the absence of random seeds for all methods with released code, none of the methods in Table~\ref{tab:reproduce} are exactly reproducible from the search phase to the final architecture evaluation phase.  

While only criteria 1--3 are necessary to estimate broad reproducibility, there is minimal discussion of the broad reproducibility of existing methods in published work.  With the exception of NASBOT \citep{Kandasamy2018} and DARTS \citep{liu2018darts}, the methods in Table~\ref{tab:reproduce} only report the performance of the best found architecture, presumably resulting from a single run of the search process.  Although this is understandable in light of the computational costs for some of these methods \citep{Luo2018, cai2018path}, the high variance of extremal statistics makes it difficult to isolate the impact of the novel contributions introduced in each work.  DARTS is  particularly commendable in acknowledging its dependence on random initialization, prompting the use multiple runs to select the best architecture.  In our experiments in Section~\ref{sec:exp}, we follow DARTS and report the result of our random weight-sharing method across multiple trials; in fact,  we go one step further and evaluate the broad reproducibility of our results with multiple sets of random seeds.

\section{Methodology}
\label{sec:alg}

We now introduce our NAS algorithm that combines random search with weight-sharing. Our algorithm is designed for an arbitrary search space with a DAG representation, and in our in our experiments in Section~\ref{sec:exp}, we use the same search spaces as that considered by DARTS \citep{liu2018darts} for the standard CIFAR-10 and PTB NAS benchmarks. 

\begin{figure}
    \centering
    \includegraphics[width=0.5\textwidth]{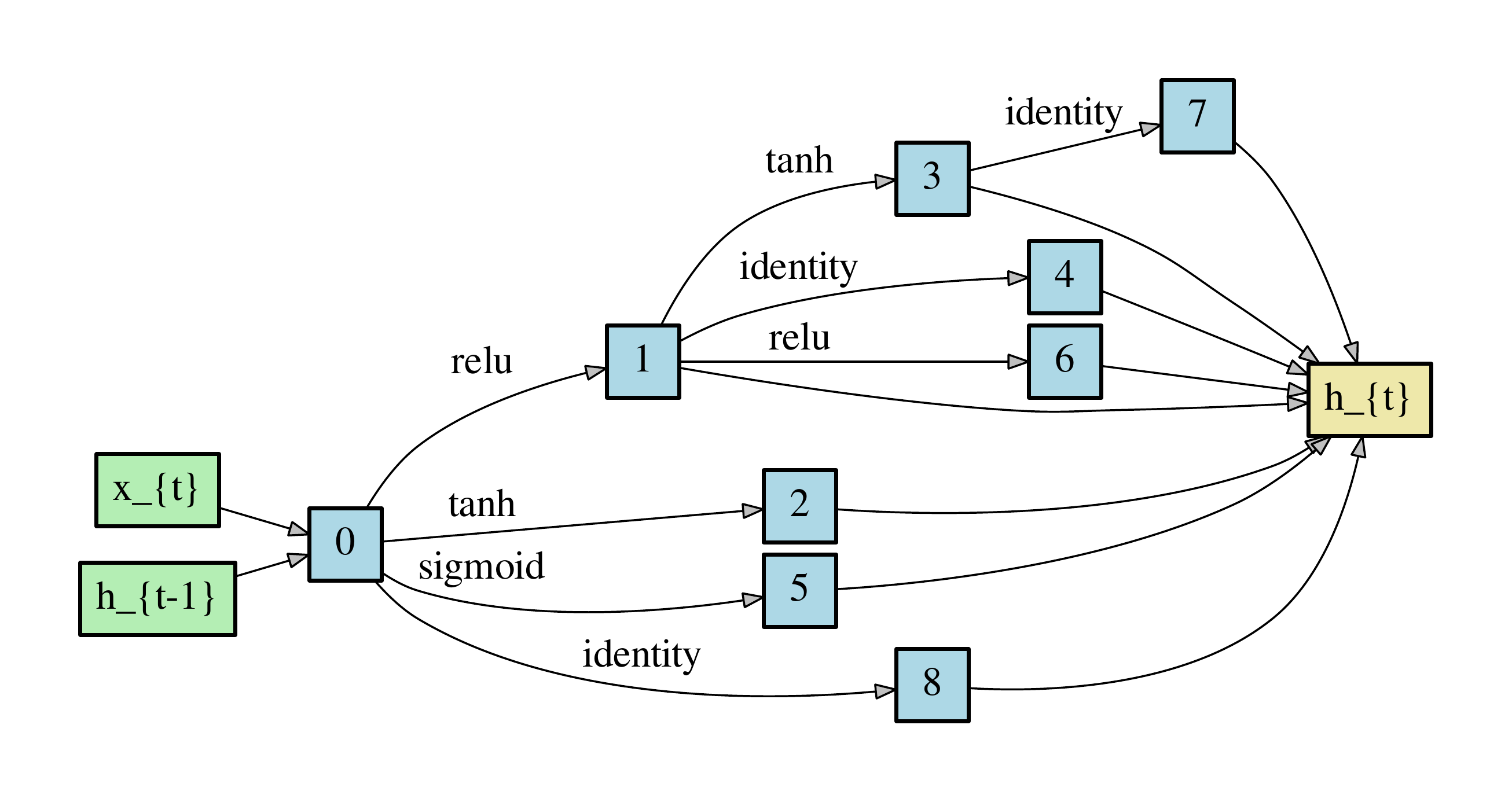}
    \caption{\textbf{Recurrent Cell on PTB Benchmark.}  The best architecture found by random search with weight-sharing in Section~\ref{ssec:rnn} is depicted. Each numbered square is a node of the DAG and each edge represents the flow of data from one node to another after applying the indicated operation along the edge.  Nodes with multiple incoming edges (i.e., node 0 and output node \texttt{h\_\{t\}} concatenate the inputs to form the output of the node.  }
    \label{fig:rnn_cell}
\end{figure}

For concreteness, consider the search space used by DARTS for designing a recurrent cell for the PTB benchmark: the DAG considered for the recurrent cell has $N=8$ nodes and the operations considered include tanh, relu, sigmoid, and identity.
To sample an architecture from this search space, we apply random search in the following manner: 

\begin{enumerate}[leftmargin=*]
    \item For each node in the DAG, determine what decisions must be made.  In the case of the PTB search space, we need to choose a node as input and a corresponding operation to apply to generate the output of the node. 
    \item For each decision, identify the possible choices for the given node. In the case of the PTB search space, if we number the nodes from $1$ to $N$, node $i$ can take the outputs of nodes $0$ to node $i-1$ as input (the initial input to the cell is index 0 and is also a possible input).   Additionally, we can choose an operation from $\{\text{tanh, relu, sigmoid, and identity}\}$ to apply to the output of node $i$.  
    \item Finally, moving from node to node, we sample uniformly from the set of possible choices for each decision that needs to be made.
\end{enumerate}
Figure~\ref{fig:rnn_cell} shows an example of an architecture from this search space.

In order to combine random search with weight-sharing, we simply use randomly sampled architectures to train the shared weights.  Shared weights are updated by selecting a single architecture for a given minibatch and updating the shared weights by back-propagating through the network with only the edges and operations as indicated by the architecture activated.  Hence, the number of architectures used to update the shared weights is equivalent to the total number of minibatch training iterations.  

After training the shared weights for a certain number of epochs, we use these trained shared weights to evaluate the performance of a number of randomly sampled architectures on a separate held out dataset. We select the best performing one as the final architecture, i.e., as the output of our search algorithm.

\subsection{Relevant Meta-Hyperparameters}
\label{ssec:relevant_hps}

There are a few  key meta-hyperparameters that impact the behavior of our search algorithm. We describe each of them below, along with a  description of how we expect them to impact the search algorithm, both in terms of search quality and computational costs. 

\begin{enumerate}
    \item \textbf{Training epochs.}  Increasing the number of training epochs while keeping all other parameters the same increases the total number of minibatch updates and hence, the number of architectures used to update the shared weights.  Intuitively, training with more architectures should help the shared weights generalize better to what are likely unseen architectures in the evaluation step.  Unsurprisingly, more epochs increase the computational time required for architecture search.
    \item \textbf{Batch size.}  Decreasing the batch size while keeping all other parameters the same also increases the number of minibatch updates but at the cost of noisier gradient update.  Hence, we expect reducing the batch size to have a similar effect as increasing the number of training epochs but may necessitate adjusting other meta-hyperparameters to account for the noisier gradient update.  Intuitively, more minibatch updates increase the computational time required for architecture search.
    \item \textbf{Network size.} Increasing the search network size increases the dimension of the shared weights.  Intuitively, this should boost performance since a larger search network can store more information about different architectures.  Unsurprisingly, larger networks require more GPU memory.
    \item \textbf{Number of evaluated architectures.}  Increasing the number of architectures that we evaluate using the shared weights allows for more exploration in the architecture search space.  Intuitively, this should help assuming that there is a high correlation between the performance of an architecture evaluated using shared weights and the ground truth performance of that architecture when trained from scratch \citep{bender2018understanding}.  Unsurprisingly, evaluating more architectures increases the computational time required for architecture search. 
\end{enumerate}
Other learning meta-hyperparameters will likely need to be adjusted accordingly for different settings of the key relevant meta-hyperparameters listed above. In our experiments in Section~\ref{sec:exp}, we tune \emph{gradient clipping} as a fifth meta-hyperparameter, though there are other possible meta-hyperparameters that may benefit from additional tuning (e.g., learning rate, momentum).

In Section~\ref{sec:exp}, following these intuitions, we incrementally explore the design space of our search method in order to improve search quality and make full use of the available GPU memory and computational resources. 

\subsection{Memory Footprint}
Since we train the shared weights using a single architecture at a time, we have the option of only loading the weights associated with the operations and edges that are activated into GPU memory.  Hence, the memory footprint of our random search with weight-sharing can be reduced to that of a single model.  In this sense, our approach is similar to ProxylessNAS \citep{cai2018proxylessnas} and allows us to perform architecture search with weight-sharing on the larger ``proxyless'' models that are usually used in the final architecture evaluation step instead of the smaller proxy models that are usually used in the search step.  We take advantage of this in a subset of our experiments for the PTB benchmark in Section~\ref{ssec:rnn}; performing random search with weight-sharing on a proxyless network for the CIFAR-10 benchmark is a direction for future work.

In contrast, \citet{bender2018understanding} trains the shared weights with a path dropout schedule that incrementally prunes edges within the DAG so that the sub-DAGs used to train the shared weights become sparser as training progresses.  Under this training routine, since most of the edges in the search DAG are activated in the beginning, the memory footprint cannot be reduced to that of a single model to allow a proxyless network for the shared weights.

\section{Experiments}
\label{sec:exp}

In line with prior work \citep{nasRL, pham18ENAS, liu2018darts}, we consider the two standard benchmarks for neural architecture search:
(1) language modeling on the Penn Treebank (PTB) dataset \citep{ptb} and (2) image classification on CIFAR-10 \citep{cifar10data}. 
For each of these benchmarks, we consider the same search space and use much of the same experimental setups as DARTS \citep{liu2018darts}, and by association SNAS \citep{xie2018snas}, to facilitate a fair comparison of our results to existing work. 

To evaluate the performance of random search with weight-sharing on these two benchmarks, we proceed in the same three stages as \citet{liu2018darts}:
\begin{itemize}
    \item \textbf{Stage 1}: Perform architecture search for a cell block on a cheaper search task.  
    \item \textbf{Stage 2}: Evaluate the best architecture from the first stage by retraining a larger, network formed from multiple cell blocks of the best found architecture from scratch.  This stage is used to select the best architecture from multiple trials.
    \item \textbf{Stage 3}: Perform the full evaluation of the best found architecture from the second stage by either training for more epochs (PTB) or training with more seeds (CIFAR-10).  
\end{itemize}
We start with the same meta-hyperparameter settings used by DARTS to train the shared weights.  Then, we incrementally modify the meta-hyperparameters identified in Section~\ref{ssec:relevant_hps} to improve performance until we either reach state-of-the-art performance (for PTB) or match the performance of DARTS and SNAS (for CIFAR-10).  

For our evaluation of random search with early-stopping (i.e., ASHA) on these two benchmarks, we perform architecture search using partial training of the stage (2) evaluation network and then select the best architecture for stage (3) evaluation.  For both benchmarks, we run ASHA with a starting resource per architecture of $r=1$ epoch, a maximum resource of $300$ epochs, and a promotion rate of $\eta=4$, indicating the top $1/4$ of architectures will be promoted in each round and trained for $4\times$ more resource. 

\subsection{PTB Benchmark}
\label{ssec:rnn}
We now present results for the PTB benchmark.  
We use the DARTS search space for the recurrent cell, which is described in Section~\ref{sec:alg}.  For this benchmark, 
due to higher memory requirements for their mixture operation, DARTS used a small recurrent network with embedding and hidden dimension of 300 to perform the architecture search followed by a larger network with embedding and hidden dimension of $850$ to perform the evaluation.  For the PTB benchmark, we refer to the network used in the first stage as the \emph{proxy} network and the network in the later stages as the \emph{proxyless} network.  
 We next present the final search results. We subsequently explore the impact of various meta-hyperparameters on random search with weight-sharing, and finally evaluate the reproducibility of various methods on this benchmark.

\subsubsection{Final Search Results}

We now present our final evaluation results in Table~\ref{tab:rnn_sota}.  Specifically, we report the output of stage (3), in which we train the proxyless network configured according to the best architectures found by different methods for 3600 epochs. This setup matches the evaluation scheme used for the reported results in Table 2 of \citet{liu2018darts} (see Appendix~\ref{ssec:appendix_reproduce} for more details).  
We discuss various aspects of these results in the context of the three issues---baselines, complex methods, reproducibility---introduced in Section~\ref{sec:intro}.

\begin{table}[h]
\caption{\textbf{PTB Benchmark: Comparison with state-of-the-art NAS methods and manually designed networks.} Lower test perplexity is better on this benchmark.  The results are grouped by those for manually designed networks, published NAS methods, and the methods that we evaluated.  Table entries denoted by "-" indicate that the field does not apply, while entries denoted by "N/A" indicate unknown entries. The search cost, unless otherwise noted, is measured in GPU days.  Note that the search cost is hardware dependent and the search cost shown for our results are calculated for Tesla P100 GPUs; all other numbers are those reported by \citet{liu2018darts}. 
\\\hspace{\textwidth}
$^\#$ Search cost is in CPU-days.
\\\hspace{\textwidth}
$^*$ We could not reproduce this result using the code released by the authors at \url{https://github.com/melodyguan/enas}.
\\\hspace{\textwidth}
$^\dagger$ The stage (1) cost shown is that for 1 trial as opposed to the cost for 4 trials shown for DARTS and Random search WS.  It is unclear whether ENAS requires multiple trials followed by stage (2) evaluation in order to find a good architecture. 
}
\label{tab:rnn_sota}
\scalebox{0.8}{
\begin{tabular}{lccccccccc}
\hline
& &\multicolumn{2}{c}{\textbf{Test Perplexity}} & \textbf{Params} & \multicolumn{3}{c}{\textbf{Search Cost}} & \textbf{Comparable} & \textbf{Search} \\
\textbf{Architecture} & \textbf{Source} & Valid & Test & (M) & Stage 1 & Stage 2 & Total & \textbf{Search Space?} & \textbf{Method} \\

\hline
LSTM + DropConnect & \citep{merity2018awd} & 60.0 & 57.3 & 24 & - & - & - & - & manual \\
ASHA + LSTM + DropConnect & \citep{asha} & 58.1 & 56.3 & 24 & - & - & 13 & N & HP-tuned\\
LSTM + MoS & \citep{yang2018breaking} & 56.5 & 54.4 & 22 & - & - & - & - & manual \\
\hline
NAS$^\#$ & \citep{nasRL} & N/A & 64.0 & 25 & - & - & 1e4 &  N & RL \\
ENAS${^*}{^\dagger}$ & \citep{pham18ENAS} & N/A & 56.3 & 24 & 0.5 & N/A & N/A & Y & RL \\
ENAS$^\dagger$ & \citep{liu2018darts} & 60.8 & 58.6 & 24 & 0.5 & N/A & N/A &  Y & random \\
 Random search baseline & \citep{liu2018darts} & 61.8 & 59.4 & 23 & - & - & 2 & Y & random \\
 DARTS (first order)& \citep{liu2018darts}     &  60.2 & 57.6 & 23 & 0.5 & 1 & 1.5 & Y & gradient-based\\
 DARTS (second order)&  \citep{liu2018darts}&  58.1 & 55.7 & 23 & 1 & 1& 2 & Y & gradient-based \\
 \hline
 DARTS (second order)& Ours & 58.2 & 55.9 & 23 & 1 & 1 & 2&Y & gradient-based \\
 ASHA baseline  & Ours & 58.6 & 56.4 & 23 & - & - & 2 & Y & random \\
 Random search WS& Ours & 57.8 & 55.5 & 23 & 0.25 & 1 & 1.25 & Y & random \\
 \hline
 
\end{tabular}}
\end{table}

First, we evaluate the ASHA baseline using 2 GPU days, which is equivalent to the total cost of DARTS (second order). 
In contrast to the one random architecture evaluated by \citet{pham18ENAS} and the 8 evaluated by \citet{liu2018darts} for their random search baselines, ASHA evaluated over 300 architectures with the allotted computation time. The best architecture found by ASHA achieves a test perplexity of 56.4, which is comparable to the published result for ENAS and significantly better than the random search baseline provided by \citet{liu2018darts}, DARTS (first order), and the reproduced result for ENAS \citep{liu2018darts}.  Our result demonstrates that the gap between SOTA NAS methods and standard hyperparameter optimization approaches on the PTB benchmark is significantly smaller than that suggested by the existing comparisons to random search \citep{pham18ENAS, liu2018darts}.  

Next, 
we evaluate random search with weight-sharing with tuned meta-hyperparameters (see Section~\ref{ssec:rnn_random} for details).
With slightly lower search cost than DARTS, this method 
finds an architecture that reaches test perplexity 55.5, achieving SOTA perplexity compared to previous NAS approaches.  
We note that manually designed architectures are competitive with RNN cells designed by NAS methods on this benchmark.  In fact, the work by \citet{yang2018breaking} using LSTM with mixture of experts in the softmax layer (MoS)  outperforms automatically designed cells.  Our architecture would likely also improve significantly with MoS, but we train without MoS to provide a fair comparison to ENAS and DARTS.  

Finally, we examine the reproducibility of the NAS methods with available code for both architecture search and evaluation. For DARTS, exact reproducibility was not feasible since~\citet{liu2018darts} do not provide random seeds for the search process; however, we were able to reproduce the performance of their reported best architecture.  We also evaluated the broad reproducibility of DARTS through an independent run, which reached a test perplexity of 55.9, 
compared to the published value of 55.7.  For ENAS, end-to-end exact reproducibility was infeasible due to non-deterministic code and missing random seeds for both the search and evaluation steps.  Additionally, when we tried to reproduce their result using the provided final architecture, we could not match the reported test perplexity of 56.3 in our rerun.  Consequently, in Table~\ref{tab:rnn_sota} we  show the test perplexity for the final architecture found by ENAS trained using the DARTS code base, which~\citet{liu2018darts} observed to give a better test perplexity than using the architecture evaluation code provided by ENAS.
We next considered the reproducibility of random search with weight-sharing. We verified the exact reproducibility of our reported results, and then investigated their broad reproducibility by running another experiment with different random seeds.  In this second experiment, we observed a final text perplexity of 56.5, compared with a final test perplexity of 55.5 in the first experiment.
Our detailed investigation in Section~\ref{ssec:rnn_reproduce} shows that the discrepancies across both DARTS and random search with weight-sharing are unsurprising in light of the differing convergence rates among architectures on this benchmark.

 \subsubsection{Impact of Meta-Hyperparameters}
 \label{ssec:rnn_random}
We now detail the meta-hyperparameter settings that we tried for random search with weight-sharing in order to achieve SOTA performance on the PTB benchmark. Similar to DARTS, in these preliminary experiments we performed 4 separate trials of each version of random search with weight-sharing, where each trial consists of executing stage (1) followed by stage (2). In stage (1), we train the shared weights and then use them to evaluate 2000 randomly sampled architectures.  In stage (2), we select the best architecture out of 2000, according to the shared weights, to train from scratch using the proxyless network for 300 epochs. 

We incrementally tune random search with weight-sharing by adjusting the following meta-hyperparameters associated with training the shared weights in stage (1): (1) gradient clipping, (2) batch size, and (3) network size.  The settings we consider proceed as follows: 
\begin{itemize}
    \item \textbf{Random (1):} We train the shared weights of the proxy network using the same setup as DARTS with the same values for number of epochs, batch size, and gradient clipping; all other meta-hyperparameters are the same.
    \item \textbf{Random (2):} We decrease the maximum gradient norm to account for discrete architectures, as opposed to the weighted combination used by DARTS, so that gradient updates are not as large in each direction.
    \item \textbf{Random (3):} We decrease batch size from 256 to 64 in order to increase the number of architectures used to train the shared weights. 
    \item \textbf{Random (4):} We train the larger proxyless network architecture with shared weights instead of the proxy network, thereby significantly increasing the number of parameters in the model.   
\end{itemize} 
The stage (2) performance of the final architecture after retraining from scratch for each of these settings is shown in Table~\ref{tab:rnn_random}.
With the extra capacity in the larger network used in Random (4), random search with weight-sharing achieves average validation perplexity of 64.7 across 4 trials, 
with the best architecture (shown in Figure~\ref{fig:rnn_cell} in Section~\ref{sec:alg}) reaching 63.8.  
In light of these stage (2) results, we focused in stage (3) on the best architecture found by Random (4) Run 1, and achieved test perplexity of 55.5 after training for 3600 epochs as reported in Table~\ref{tab:rnn_sota}.

\begin{table}[h]
    \centering
    \caption{\textbf{PTB Benchmark: Comparison of Stage (2) Intermediate Search Results for Weight-Sharing Methods.}  In stage (1), random search is run with different settings to train the shared weights.  The resulting networks are used to evaluate 2000 randomly sampled architectures.  In stage (2), the best of these architectures for each trial is then trained from scratch for 300 epochs.  We report the performance of the best architecture after stage (2) across 4 trials for each search method.   }
    \label{tab:rnn_random}
    \scalebox{0.8}{
    \begin{tabular}{lcccc|cccccc}
       \hline
    \multicolumn{5}{c|}{\textbf{Setting}} & &&&\\
     & Network  &  & Batch  & Gradient  &  \multicolumn{6}{c}{\textbf{Trial}}  \\
    Method &  Config & Epochs &  Size &  Clipping &  1& 2 &3  &4  & Best & Average \\
    \hline
    DARTS \citep{liu2018darts} &     proxy & 50 & 256 & 0.25  & 67.3 & 66.3 & 63.4  & 63.4 & \textbf{63.4} & \textbf{65.1} \\
    Reproduced DARTS &     proxy & 50 & 256 & 0.25  & 64.5 & 67.7 & 64.0  & 67.7 & 64.0 & 66.0 \\
    \hline
    Random (1) & proxy & 50 & 256 & 0.25  & 65.6 & 66.3 & 66.0 & 65.6 & 65.6  & 65.9\\
     Random (2) & proxy & 50 & 256 & 0.1 & 65.8 & 67.7 & 65.3 & 64.9 & 64.9 & 65.9 \\
         Random (3) & proxy & 50 & 64 & 0.1 & 66.1 & 65.0 & 64.9 & 64.5 & 64.5 & 65.1 \\
         Random (4) Run 1 & proxyless & 50 & 64 & 0.1 & 66.3 & 64.6 & 64.1 & 63.8 & \textbf{63.8}& \textbf{64.7} \\
         \hline
         Random (4) Run 2 & proxyless & 50 & 64 & 0.1 & 63.9 & 64.8 & 66.3 & 66.7 & 63.9 & 65.4 \\
         \hline
    \end{tabular}
    }
    
\end{table}

\subsubsection{Investigating Reproducibility}
\label{ssec:rnn_reproduce}
\begin{table}[]
    \centering
    \caption{\textbf{PTB Benchmark: Ranking of Intermediate Validation Perplexity.}  Architectures are retrained from scratch using the proxyless network and the validation perplexity is reported after training for the indicated number of epochs.  The final test perplexity after training for 3600 epochs is also shown for reference.}
    \scalebox{0.8}{\begin{tabular}{c|cccccccccc|cc}
    \hline
     & \multicolumn{10}{c|}{\textbf{Validation Perplexity by Epoch}} & \multicolumn{2}{c}{\textbf{Test}}\\
     & \multicolumn{2}{c}{300} & \multicolumn{2}{c}{500} & \multicolumn{2}{c}{1600} & \multicolumn{2}{c}{2600} &  \multicolumn{2}{c|}{3600} & \multicolumn{2}{c}{\textbf{Perplexity}}\\
    \textbf{Search Method } & Value & Rank & Value & Rank & Value & Rank & Value & Rank & Value & Rank & Value & Rank\\
    \hline
         DARTS & 64.0 & 4 & 61.9 & 2& 59.5 & 2 & 58.5 & 2 & 58.2 & 2 & 55.9 &2\\
         ASHA & 63.9 & 2 & 62.0 & 3 & 59.8 & 4 & 59.0 &  3 & 58.6 & 3 & 56.4 & 3 \\
         Random (4) Run 1 & 63.8 & 1 & 61.7 &1 & 59.3 & 1 & 58.4 & 1 & 57.8 & 1 & 55.5&1\\
         Random (4) Run 2 & 63.9 & 2 & 62.1 & 4 & 59.6 & 3 & 59.0 & 3 & 58.8 & 4 & 56.5&4\\
         \hline
    \end{tabular}}
    
    \label{tab:rnn_reproduce}
\end{table}

We next examine the  stage (2) intermediate results in Table~\ref{tab:rnn_random} in the context of reproducibility. 
The first two rows of Table~\ref{tab:rnn_random} show a comparison of the published stage (2) results for DARTS and our independent runs of DARTS. 
Both the best and average across 4 trials are worse in our reproduction of their results. 
Additionally, as previously mentioned, we perform an additional run of Random (4) with 4 different random seeds to test the broad reproducibility our result. The minimum stage (2) validation perplexity over these 4 trials 
is 63.9, compared to a minimum validation perplexity of 63.8 for the first set of seeds.  

Next, in Table~\ref{tab:rnn_reproduce} we compare the validation perplexities of 
the best architectures from ASHA, Random (4) Run 1, Random (4) Run 2, and our independent run of DARTS after
training each from scratch for up to 3600 epochs.  The swap in relative ranking across epochs demonstrates the risk of using noisy signals for the reward. In this case, we see that even partial training for 300 epochs does not recover the correct ranking; training using shared weights further obscures the signal. The differing convergence rates explain the difference in final test perplexity of the best architecture from Random (4) Run 2 and those from DARTS and Random (4) Run 1, despite Random (4) Run 2 reaching a comparable perplexity after 300 epochs. 

Overall, the results of Tables~\ref{tab:rnn_random} and~\ref{tab:rnn_reproduce} demonstrate a high variance in the stage (2) intermediate results across trials, along with issues related to differing convergence rates for different architectures. These two issues help explain the differences between the independent runs of DARTS and random search with weight-sharing.  A third potential source of variation, which could in particular adversely impact our random search with weight-sharing results, stems from the fact that we did not perform any additional hyperparameter tuning in stage (3); instead we used the same training hyperparameters that were tuned by \citet{liu2018darts} for the final architecture found by DARTS.

\subsection{CIFAR-10 Benchmark}
\label{ssec:cnn}
We next present results for the CIFAR-10 benchmark. The DAG considered for the convolutional cell has $N=4$ search nodes and the operations considered include $3\times 3$ and $5\times 5$ separable convolutions, $3\times 3$ and $5\times 5$ dilated separable convolutions, $3\times 3$ max pooling, and $3\times 3$ average pooling, and zero \citep{liu2018darts}.  To sample an architecture from this search space, we have to choose, for each node, 2 input nodes from previous nodes and associated operations to perform on each input (there are two initial inputs to the cell that are also possible input); we sample in this fashion twice, once for the normal convolution cell and one for the reduction cell (e.g., see Figure~\ref{fig:cnn_cell}).  Note that in contrast to DARTS, we include the zero operation when choosing the final architecture for each trial for further evaluation in stages (2) and (3).  We hypothesize that our results may improve if we impose a higher complexity on the final architectures by excluding the zero op.  

Due to higher memory requirements for weight-sharing, \citet{liu2018darts} uses a smaller network with 8 stacked cells and 16 initial channels to perform the convolutional cell search, followed by a larger network with 20 stacked cells and 36 initial channels to perform the evaluation.  Again, we will refer to the network used in the first stage as the proxy network and the network in the second stage the proxyless network.  

Similar to the PTB results in Section~\ref{ssec:rnn}, we will next present the final search results for the CIFAR-10 benchmark, and then dive deeper into these results to explore the impact of meta-hyperparameters on stage (2) intermediate results, and finally evaluate associated reproducibility ramifications.

\subsubsection{Final Search Results}

\begin{table}[h]
\caption{\textbf{CIFAR-10 Benchmark: Comparison with state-of-the-art NAS methods and manually designed networks.}  The results are grouped by those for manually designed networks, published NAS methods, and the methods that we evaluated.  Models for all methods are trained with auxiliary towers and cutout. Test error for our contributions are averaged over 10 random seeds. Table entries denoted by "-" indicate that the field does not apply, while entries denoted by "N/A" indicate unknown entries. The search cost is measured in GPU days.  Note that the search cost is hardware dependent and the search cost shown for our results are calculated for Tesla P100 GPUs; all other numbers follow those reported by \citet{liu2018darts}.
\\\hspace{\textwidth}
$^*$ We show results for the variants of these networks with comparable number of parameters.  Larger versions of these networks achieve lower errors.
\\\hspace{\textwidth}
$^\#$ Reported test error averaged over 5 seeds.
\\\hspace{\textwidth}
$^\dagger$ The stage (1) cost shown is that for 1 trial as opposed to the cost for 4 trials shown for DARTS and Random search WS.  It is unclear whether the method requires multiple trials followed by stage (2) evaluation in order to find a good architecture. 
\\\hspace{\textwidth}
$^\ddagger$ Due to the longer evaluation we employ in stage (2) to account for unstable rankings, the cost for stage (2) is 1 GPU day for results reported by \citet{liu2018darts} and 6 GPU days for our results.
}
\label{tab:cnn_sota}
\scalebox{0.8}{\begin{tabular}{lccccccccc}
\hline
& & \multicolumn{2}{c}{\textbf{Test Error} \hspace{0.5cm} }& \textbf{Params}  & \multicolumn{3}{c}{\textbf{Search Cost}} & \textbf{Comparable} & \textbf{Search} \\ 
\textbf{Architecture} & \textbf{Source} & Best & Average & (M) & Stage 1 & Stage 2 & Total & \textbf{Search Space?} & \textbf{Method} \\
\hline
Shake-Shake$^\#$ & \citep{cutout} & N/A & $2.56$ & 26.2 & - & - & - & - & manual \\
PyramidNet & \citep{shakedrop} & 2.31 & N/A & 26 &- &- &- & - & manual \\
\hline
NASNet-A${^\#}{^*}$ & \citep{Zoph2018LearningTA} & N/A & 2.65 & 3.3 &- &-&2000 & N & RL \\
AmoebaNet-B$^*$ & \citep{Real2018} & N/A & $2.55 \pm 0.05$ & 2.8 &- &-&3150 & N & evolution \\
ProxylessNAS$^\dagger$ & \citep{cai2018proxylessnas} & 2.08 & N/A & 5.7 & 4 & N/A & N/A & N & gradient-based \\
GHN${^\#}{^\dagger}$ & \citep{zhang2018graph} & N/A & $2.84 \pm 0.07$ & 5.7 & 0.84 & N/A & N/A & N & hypernetwork \\
SNAS$^\dagger$ & \citep{xie2018snas}  & N/A & $2.85 \pm 0.02$ & 2.8 & 1.5& N/A & N/A& Y & gradient-based \\
ENAS$^\dagger$ & \citep{pham18ENAS} & 2.89 & N/A & 4.6 & 0.5 & N/A & N/A & Y & RL \\
ENAS & \citep{liu2018darts} & 2.91 & N/A & 4.2 & 4 & 2& 6 &Y & RL \\
 Random search baseline & \citep{liu2018darts} & N/A & $3.29 \pm 0.15$ & 3.2 & -&-&4 & Y & random \\
 DARTS (first order)   & \citep{liu2018darts}   &  N/A & $3.00 \pm 0.14$ & 3.3 & 1.5 & 1 & 2.5 & Y & gradient-based\\
 DARTS (second order)& \citep{liu2018darts} &  N/A & $2.76 \pm 0.09$ & 3.3 & 4 & 1& 5 &Y & gradient-based \\
\hline
DARTS (second order)$^\ddagger$&  Ours &  $2.62$ & $2.78\pm 0.12$ & 3.3 & 4 & 6 & 10 & Y & gradient-based \\
ASHA baseline & Ours & 2.85 & $3.03\pm 0.13$& 2.2 & - & - &9 & Y & random\\
 Random search WS$^\ddagger$ &  Ours & 2.71 & $2.85\pm 0.08 $& 4.3 & 2.7 & 6 & 9.7 &Y & random \\
 \hline
 
\end{tabular}
}

\end{table}
We now present our results after performing the final evaluation in stage (3).  We use the same evaluation scheme used to produce the results in Table 1 of \citet{liu2018darts}.  In particular, we train the proxyless network configured according to the best architectures found by different methods with 10 different seeds and report the average and standard deviation.  Again, we discuss our results in the context of the three issues introduced in Section~\ref{sec:intro}.

First, we evaluate the ASHA baseline using 9 GPU days, which is comparable to the 10 GPU days we allotted to our independent run of DARTS. 
In contrast to the one random architecture evaluated by \citet{pham18ENAS} and the 24 evaluated by \citet{liu2018darts} for their random search baselines, ASHA evaluated over 700 architectures in the allotted computation time.
The best architecture found by ASHA achieves an average error of $3.03\pm 0.13$, which is significantly better than the random search baseline provided by \citet{liu2018darts} and comparable to DARTS (first order).  Additionally, the best performing seed reached a test error of 2.85, which is lower than the published result for ENAS.   
Similar to the PTB benchmark, these results suggest that the 
gap between SOTA NAS methods and standard hyperparameter optimization is much smaller than previously reported~\citep{pham18ENAS, liu2018darts}. 

Next, we evaluate random search with weight-sharing with tuned meta-hyperparameters (see Section~\ref{ssec:cnn_search} for details).
This method finds an architecture that achieves an average test error of $2.85\pm 0.08$, which is comparable to the reported results for SNAS and DARTS, the top 2 weight-sharing algorithms that use a comparable search space, as well as GHN \citep{zhang2018graph}.  Note that while the two manually tuned architectures we show in Table~\ref{tab:cnn_sota} outperform the best architecture discovered by random search with weight-sharing, they have over $7\times$ more parameters.  Additionally, the best-performing efficient NAS method, ProxylessNAS, uses a larger proxyless network and a significantly different search space than the one we consider.
As mentioned in Section~\ref{sec:alg}, random search with weight-sharing can also directly search over larger proxyless networks since it trains using discrete architectures. We hypothesize that using a proxyless network and applying random search with weight-sharing to the same search space as ProxylessNAS would further improve our results; we leave this as a direction for future work. 

Finally, we examine the reproducibility of the NAS methods using a comparable search space with available code for both architecture search and evaluation (i.e., DARTS and ENAS; to our knowledge, code is not currently available for SNAS).
For DARTS, exact reproducibility was not feasible since the code is non-deterministic and ~\citet{liu2018darts} do not provide random seeds for the search process; hence, we focus on broad reproducibility of the results.  In our independent run, DARTS reached an average test error of $2.78\pm 0.12$ compared to the published result of $2.76\pm 0.09$.  Notably, we observed that the process of selecting the best architecture in stage (2) is unstable when training stage (2) models for only 100 epochs; see Section~\ref{ssec:cnn_reproduce} for details. Hence, we use 600 epochs in all of our CIFAR experiments, including our independent DARTS run, which explains the discrepancy in stage (2) costs between original DARTS and our independent run.
For ENAS, the published results do not satisfy exact reproducibility due to the same issues as those for DARTS.  
We show in Table~\ref{tab:cnn_sota} the broad reproducibility experiment conducted by \citet{liu2018darts} for ENAS; here, ENAS found an architecture that achieved a comparable test error of 2.91 in 8$\times$ the reported stage (1) search cost.
As with the PTB benchmark, we then investigated the reproducibility of random search with weight-sharing. 
We verified exact reproducibility and then examined broad reproducibility by evaluating 5 additional independent runs of our method. We observe performance below 2.90 test error in 2 of the 5 runs and an average of 2.92 across all 6 runs. We investigate various sources for these discrepancies in Section~\ref{ssec:cnn_reproduce}.  

\begin{table}[h]
    \centering
     \caption{\textbf{CIFAR-10 Benchmark: Comparison of Stage (2) Intermediate Search Results for Weight-Sharing Methods.}  In stage (1), random search is run with different settings to train the shared weights.  The shared weights are then used to evaluate the indicated number of randomly sampled architectures.  In stage (2), the best of these architectures for each trial is then trained from scratch for 600 epochs.  We report the performance of the best architecture after stage (2) for each trial for each search method.   \\\hspace{\textwidth}
    $^\dagger$ This run was performed using the DARTS code before we corrected for non-determinism (see Appendix~\ref{ssec:appendix_reproduce}).  }
    \label{tab:cnn_random}
    \scalebox{0.8}{
         
       \begin{tabular}{ccccc|cccccc}
    \hline
    \multicolumn{5}{c|}{\textbf{Setting}} &&&&&\\
     &   & Gradient  & Initial & \# Archs  & \multicolumn{6}{c}{\textbf{Trial}} \\
     Method &   Epochs &  Clipping & Channels & Evaluated & 1 & 2 & 3 & 4 & Best & Average \\
    \hline
    Reproduced DARTS$^\dagger$ &      50  & 5 & 16 & - & 2.92 & 2.77 & 3.00 & 3.05 & \textbf{2.77} & \textbf{2.94} \\
    \hline
    Random (1) &  50 & 5  & 16 & 1000 &3.25&4.00&2.98&3.58&2.98&3.45\\
    Random (2) &  150 & 5 & 16 & 5000 & 2.93 & 3.80 & 3.19 & 2.96 & 2.93 & 3.22 \\
        Random (3) & 150  & 1 & 16 & 5000 & 3.50 & 3.42 & 2.97 & 2.95 & 2.97 & 3.21\\
         Random (4)  &  300  & 1 & 16 & 11000 & 3.04 & 2.90 &3.14& 3.09&2.90& 3.04\\
         Random (5) Run 1 & 150 & 1 & 24 & 5000 & 2.96 &3.33&2.83&3.00&\textbf{2.83}& \textbf{3.03}\\
         \hline
         
         \hline
    \end{tabular}}
   
\end{table}

\subsubsection{Impact of Meta-Hyperparameters}
\label{ssec:cnn_search}
We next detail the meta-hyperparameter settings that we tried in order to reach competitive performance on the CIFAR-10 benchmark via random search with weight-sharing.
Similar to DARTS, and as with the PTB benchmark, in these preliminary experiments we performed 4 separate trials of each version of random search with weight-sharing, where each trial consists of executing stage (1) followed by stage (2). In stage (1), we train the shared weights and use them to evaluate a given number of randomly sampled architectures on the test set.  In stage (2), we select the best architecture, according to the shared weights, to train from scratch using the proxyless network for 600 epochs. 

\begin{figure}[t]
    \centering
    \begin{subfigure}[b]{0.45\textwidth}
        \includegraphics[width=\textwidth]{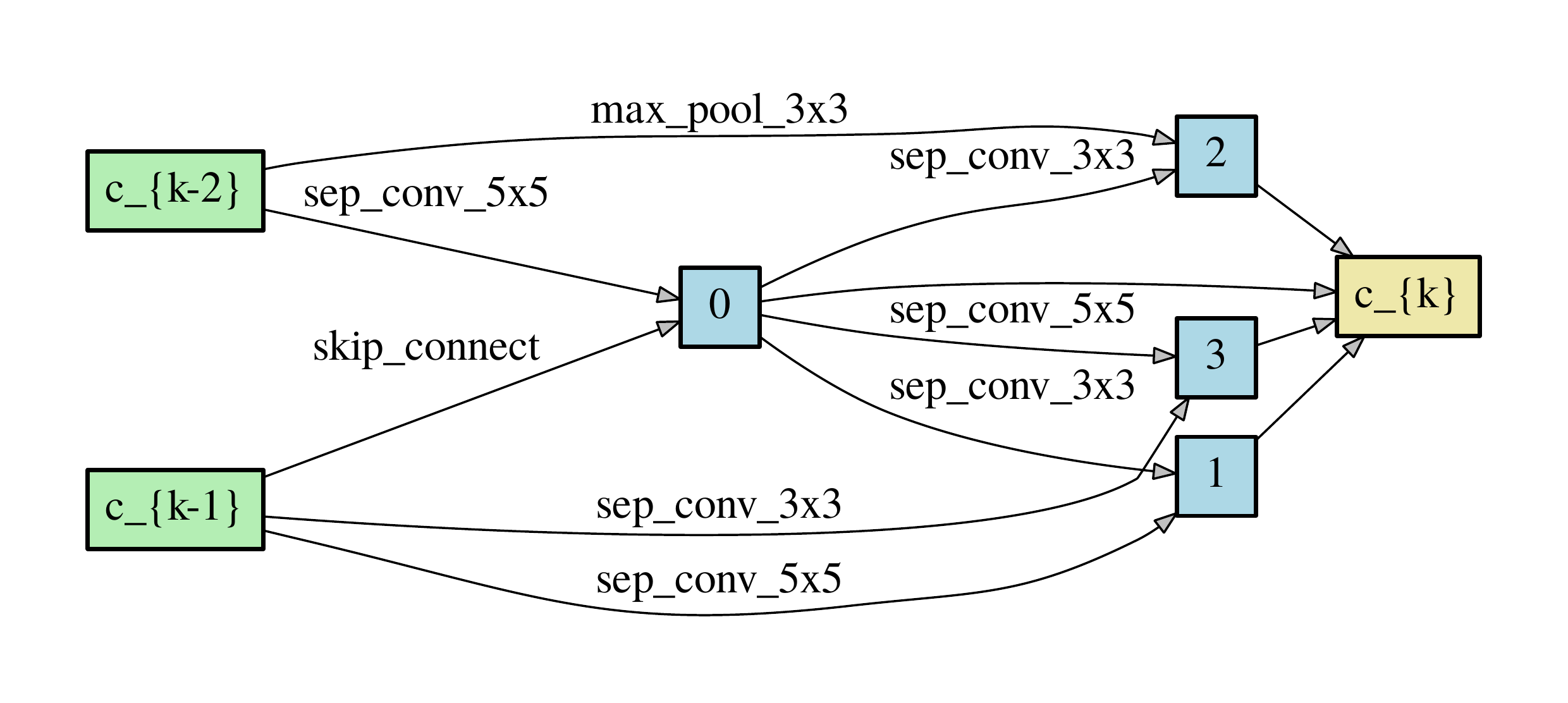}
        \caption{Normal Cell}
    \end{subfigure}
    \begin{subfigure}[b]{0.45\textwidth}
        \includegraphics[width=\textwidth]{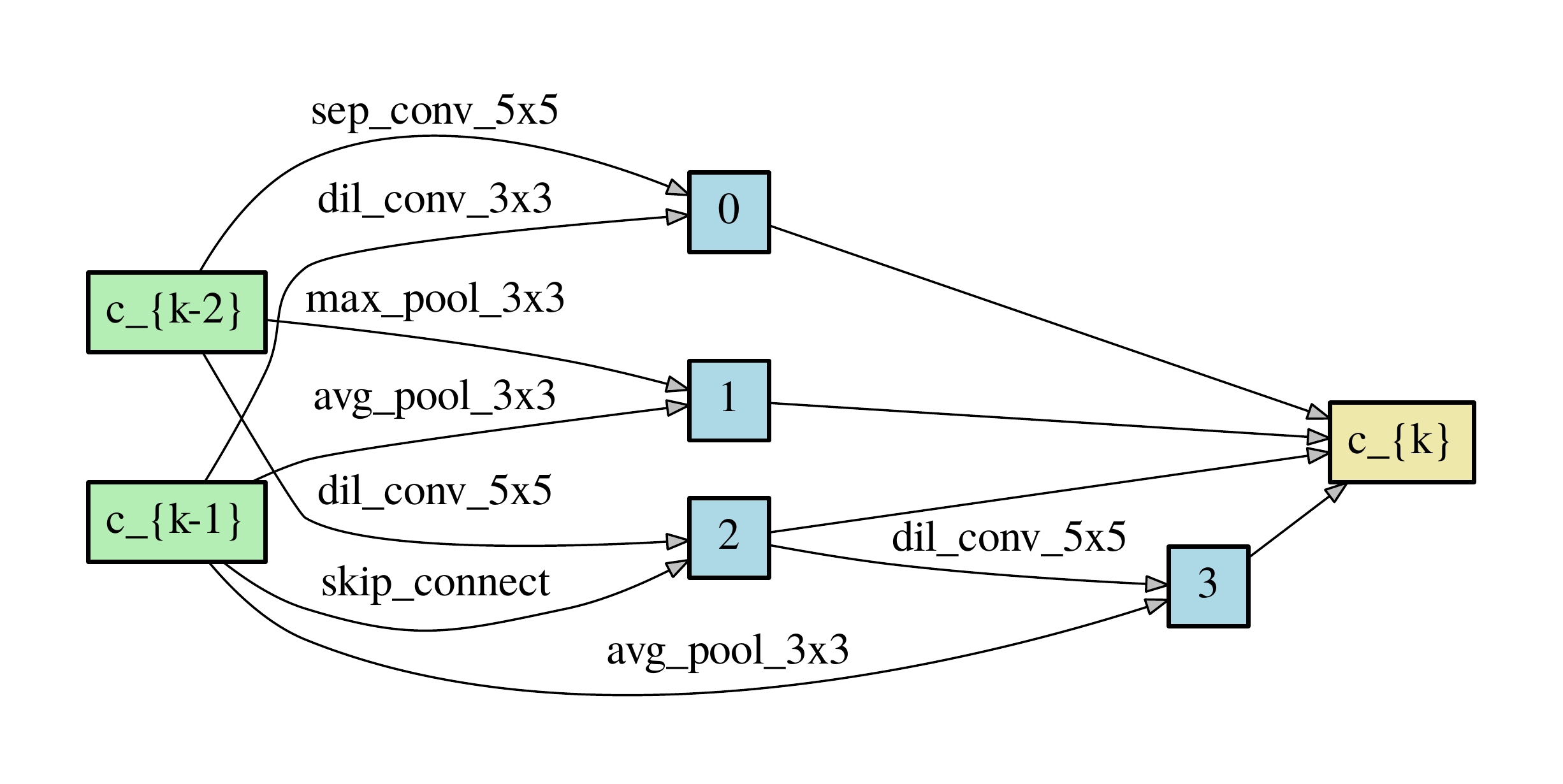}
        \caption{Reduction Cell}
    \end{subfigure}
    \caption{\textbf{Convolutional Cells on CIFAR-10 Benchmark:} Best architecture found by random search with weight-sharing.}
    \label{fig:cnn_cell}
\end{figure}

We incrementally tune random search with weight-sharing by adjusting the following meta-hyperparameters that impact both the training of shared weights and the evaluation of architectures using these trained weights: number of training epochs, gradient clipping, number of architectures evaluated using shared weights, and network size. The settings we consider for random search proceed as follows:
\begin{itemize}
    \item \textbf{Random (1):} We start by training the shared weights with the proxy network used by DARTS and default values for number of epochs, gradient clipping, and number of initial filters; all other meta-hyperparameters are the same.
    \item \textbf{Random (2):} We increase the number of training epochs from 50 to 150, which concurrently increases the number of architectures used to update the shared weights.
    \item \textbf{Random (3):} We reduce the maximum gradient norm from 5 to 1 to adjust for discrete architectures instead of the weighted combination used by DARTS.
    \item \textbf{Random (4):} We further increase the number of epochs for training the proxy network with shared weights to 300 and increase the number of architectures evaluated using the shared weights to 11k. 
    \item \textbf{Random (5):} We separately increase the proxy network size to be as large as possible given the available memory on a Nvidia Tesla P100 GPU (i.e. by $\approx 50\%$ due to increasing the number of initial channels from 16 to 24).
\end{itemize} 
The performance of the final architecture after retraining from scratch for each of these settings is shown in Table~\ref{tab:cnn_random}. 
Similar to the PTB benchmark, the best setting for random search was Random (5), which has a larger network size.  The best trial for this setting reached a test error of 2.83 when retraining from scratch; we show the normal and reduction cells found by this trial in Figure~\ref{fig:cnn_cell}.
In light of these stage (2) results, we focus in stage (3) on the best architecture found by Random (5) Run 1, and achieve an average test error of $2.85\pm 0.08$ over 10 random seeds as shown in Table~\ref{tab:cnn_sota}.

\subsubsection{Investigating Reproducibility}
\label{ssec:cnn_reproduce}
Our results in this section show that although DARTS appears broadly reproducible, this result is surprising given the unstable ranking in architectures observed between 100 and 600 epochs for stage (2) evaluation.
To begin, the first row of Table~\ref{tab:cnn_random} shows our reproduced results for DARTS after training the best architecture for each trial from scratch for 600 epochs.  In our reproduced run, DARTS reaches an average test error of 2.94 and a minimum of 2.77 across 4 trials (see Table~\ref{tab:cnn_random}).  Note that this is not a direct comparison to the published result for DARTS since there, the stage (2) evaluation was performed after training for only 100 epochs.  
\begingroup
\setlength{\tabcolsep}{3pt}
\begin{table}[h]
    \centering
    \caption{\textbf{CIFAR-10 Benchmark: Ranking of Intermediate Test Error for DARTS.}  Architectures are retrained from scratch using the proxyless network and the error on the test set is reported after training for the indicated number of epochs.  Rank is calculated across the 4 trials.  We also show the average over 10 seeds for the best architecture from the top trial for reference.\\\hspace{\textwidth}
    $^\dagger$ These results were run before we fixed the non-determinism in DARTS code (see Appendix~\ref{ssec:appendix_reproduce}).}
    
    \scalebox{0.8}{\begin{tabular}{cc|cccc|cc}
    \hline
    && \multicolumn{4}{c|}{\textbf{Epochs}}&\multicolumn{2}{c}{\textbf{Across}}\\
    \textbf{Search}&& \multicolumn{2}{c}{100} & \multicolumn{2}{c|}{600} & \multicolumn{2}{c}{\textbf{10 Seeds}}\\
         \textbf{Method} & \textbf{Trial} & Value & Rank & Value & Rank & \multicolumn{1}{c}{Min}& \multicolumn{1}{c}{Avg}\\
         \hline
         Reproduced  & 1& 7.63 & 2 & 2.92  & 2 \\
		Darts$^\dagger$ & 2 & 7.67 & 3 & 2.77 & 1& 2.62 & $2.78\pm 0.12$ \\
         &3& 8.38 & 4 & 3.00 & 3 \\
         &4& 7.51 & 1 & 3.05 & 4 \\
         \hline
         
    \end{tabular}}
    \label{tab:cnn_reproduce}
\end{table}
\endgroup

\begin{table*}[h]
    \centering
     \caption{\textbf{CIFAR-10 Benchmark: Broad Reproducibility of Random Search WS}  We report the  stage 3 performance of the final architecture from 6 independent runs of random search with weight-sharing.  \\\hspace{\textwidth}
 $^\dagger$ This run was performed using the DARTS code before we corrected for non-determinism (see Appendix~\ref{ssec:appendix_reproduce}).  }
    \label{tab:cnn_random_broad}
    \scalebox{0.8}{
  
       \begin{tabular}{cccccc|c}
    \hline
   \multicolumn{6}{c|}{\textbf{Test Error Across 10 Seeds}} & \\
Run 1 & Run $2^\dagger$ & Run 3 & Run 4 & Run 5 & Run 6 & \textbf{Average}\\
         \hline
         $2.85 \pm 0.08 $ & $2.86 \pm 0.09$ & $2.88 \pm 0.10$ & $2.95 \pm 0.09$ &$2.98 \pm 0.12$ & $3.00 \pm 0.19$ & 2.92\\
         \hline
    \end{tabular}}
   
\end{table*}
Delving into the intermediate results, we compare the performance of the best architectures across trials from our independent run of DARTS  after training each from scratch for 100 epochs and 600 epochs (see Table~\ref{tab:cnn_reproduce}).    We see that the ranking is unstable between 100 epochs and 600 epochs, which motivated our strategy of training the final architectures across trials to 600 epochs in order to select the best architecture for final evaluation across 10 seeds.  This suggests we should be cautious when using noisy signals for the performance of different architectures, especially since architecture search is conducted for DARTS and Random (5) for only 50 and 150 epochs respectively.

Finally, we investigate the variance of random search with weight-sharing with 5 additional runs as shown in  Table~\ref{tab:cnn_random_broad}.  The stage (3) evaluation of the best architecture for these 5 additional runs reveal that 2 out of 5 achieve similar performance as Run 1, while the 3 remainder underperform but still reach a better test error than ASHA.  These broad reproducibility results show that random search with weight-sharing has high variance between runs, which is not surprising given the change in intermediate rankings that we observed for DARTS.   

\subsection{Computational Cost}  
As mentioned in Section~\ref{sec:related}, there is a trade off between computational cost and the quality of the signal that we get per architecture that we evaluate.  
To get a better sense of the tradeoff, we estimate the per architecture computational cost of different methods considered in our experiments, noting that these methods differ in per architecture cost by at least an order-of-magnitude as we move from expensive to cheaper evaluation methods:
\begin{enumerate}
    \item \textbf{Full training:} The random search baselines considered by \citet{liu2018darts} cost 0.5 GPU days per architecture for the PTB benchmark and 4 GPU hours per architecture for the CIFAR-10 benchmark.
    \item \textbf{Partial training:} The amortized cost of ASHA is 9 minutes per architecture for the PTB benchmark and 19 minutes per architecture for the CIFAR-10 benchmark.
    \item \textbf{Weight-sharing:}  It is difficult to quantify the equivalent number of architectures evaluated by DARTS and random search with weight-sharing. For DARTS, the final architecture is taken to be the highest weighted operation for each architectural decision, but it is unclear how much information is provided by the gradient updates to the architecture mixture weights during architecture search.  For random search with weight sharing, although we evaluate a given number of architectures using the shared weights, as stated in Section~\ref{sec:alg}, this is a tunable meta-hyperparameter and the quality of the performance estimates we receive can be noisy.\footnote{In principle, the equivalent number of architectures evaluated can be calculated by applying an oracle CDF of ground truth performance over randomly sampled architectures to the performance of the architectures found by the shared weights.}  Nonetheless, to provide a rough estimate of the cost per architecture for random search with weight-sharing, we calculate the amortized cost by dividing the total search cost by the number of architectures evaluated using the shared weights.  Hence, the amortized cost for random search with weight-sharing is 0.2 minutes per architecture for the PTB benchmark and 0.8 minutes per architecture for the CIFAR-10 benchmark. 
\end{enumerate}

Despite the apparent computational savings from weight-sharing methods, without more robustness and transparency, it is difficult to ascertain whether the total cost of applying existing weight-sharing methods to NAS problems warrants their broad application.  In particular, the total costs of ProxylessNAS, ENAS, and SNAS are likely much higher than that reported in Table~\ref{tab:rnn_sota} and Table~\ref{tab:cnn_sota} since, as we saw with DARTS and random search with weight-sharing, multiple trials are needed due to sensitivity to initialization. Additionally, while we lightly tuned the meta-hyperparameter settings, we used DARTS' settings as our starting point and it is unclear whether the settings they use required considerable tuning.  In contrast, we were able to achieve nearly competitive performance with the default settings of ASHA using roughly the same total computation as that needed by DARTS and random search with weight-sharing.

\subsection{Available Code}
Unless otherwise noted, our results are exactly reproducible from architecture search to final evaluation using the code available at \url{https://github.com/liamcli/randomNAS_release}.  The code we use for random search with weight-sharing on both benchmarks is deterministic conditioned on a fixed random seed.  We provide the final architectures used for each of the trials shown in the tables above, as well as the random seeds used to find those architectures.  In addition, we perform no additional hyperparameter tuning for final architectures and only tune the meta-hyperparameters according to the discussion in the text itself. We also provide code, final architectures, and random seeds used for our experiments using ASHA.  However, we note that there is one uncontrolled source of randomness in our ASHA experiments---in the distributed setting, the asynchronous nature of the algorithm means that the results depend on the order in which different architectures finish (partially) training. Lastly, our experiments were conducted using Tesla P100 and V100 GPUs on Google Cloud.  We convert GPU time on V100 to equivalent time on P100 by applying a multiple of $1.5$.

\section{Conclusion}

We conclude by summarizing our results and proposing suggestions to push the field forward and foster broader adoption of NAS methods.
\begin{enumerate}
    \item \textbf{Better baselines that accurately quantify the performance gains of NAS methods.} The performance of random search with early-stopping evaluated in Section~\ref{sec:exp} reveals a surprisingly small performance gap between leading general-purpose hyperparameter optimization methods and specialized methods tailored for NAS.  In traditional hyperparameter optimization benchmarks, random search has also been shown to be a difficult baseline to beat. For these benchmarks, an informative measure of the performance of a novel algorithm is its `multiple of random search,' i.e., how much more compute would random search need to achieve similar performance~\cite{hyperband}.  
    An analogous baseline could be useful for NAS, where the impact of a novel NAS method can be quantified in terms of a multiplicative speedup relative to a standard hyperparameter optimization method such as random search with early-stopping.
    \item \textbf{Ablation studies that isolate the impact of individual NAS components.} 
    Our head-to-head experimental evaluation of two variants of random search (with early stopping and with weight-sharing) 
    allows us to pinpoint
    the performance gains associated with the cheaper weight-sharing 
    evaluation scheme. In contrast, the fact that random search with weight-sharing is comparable in performance to leading NAS methods calls into question the necessity of the auxiliary network used by GHN and the complicated algorithmic components employed by ENAS, SNAS, and DARTS.  Relatedly, while ProxylessNAS achieves better average test error on CIFAR-10 than random search with weight-sharing, it is unclear to what degree these performance gains are attributable to the search space, search method, and/or proxyless shared-weights evaluation method.  To promote scientific progress, we believe that ablation studies should be conducted to answer these questions in isolation.
  
    \item \textbf{Reproducible results that engender confidence and foster  scientific progress.} Reproducibility is a core tenet of scientific progress and crucial to promoting wider adoption of NAS methods. In traditional hyperparameter optimization, it is standard for empirical results to be reported over 10 independent experimental runs \citep{feurer2015efficient,fabolas2016,kandasamy2017cont,hyperband}.  In contrast, as we discuss Section~\ref{sec:related}, results for NAS methods are often reported over a single experimental run \citep{pham18ENAS,cai2018proxylessnas,xie2018snas,zhang2018graph}, without exact reproducibility.  This is a consequence of the steep time and computational cost required to perform NAS experiments, e.g., generating the experiments reported in this paper alone required several months of wall-clock time and tens of thousands of dollars. However, in order to adequately differentiate between various methods, results need to be reported over several independent experimental runs, especially given the nature of the extremal statistics that are being reported. Consequently, we conclude that either significantly more computational resources need to be devoted to evaluating NAS methods and/or more computationally tractable benchmarks need to be developed to lower the barrier for performing adequate empirical evaluations.  Relatedly, we feel it is imperative to evaluate the merits of NAS methods not only on their accuracy, but also on their robustness across these independent runs.
    
\end{enumerate}

 \section*{Acknowledgments}
 We thank Maruan Al-Shedivat, Sebastian Caldas, Greg Ganger, Kevin Jamieson, Angela Jiang, Mikhail Khodak, Gregory Plumb, Afshin Rostamizadeh, Virginia Smith, and Daniel Wong for helpful comments and valuable discussion.  
Thanks also to Julien Siems and Frank Hutter's group for their efforts to reproduce our work, which led to insights on reproducibility and motivated additional experiments.
This work was supported in part by DARPA FA875017C0141, the National Science Foundation grants IIS1705121 and IIS1838017, an Okawa Grant, a Google Faculty Award, an Amazon Web Services Award, and a Carnegie Bosch Institute Research Award. Any opinions, findings and conclusions or recommendations expressed in this material are those of the author(s) and do not necessarily reflect the views of DARPA, the National Science Foundation, or any other funding agency.
 
\appendix
\section{Appendix}
\subsection{PTB Benchmark}
In this section, we provide additional detail for the experiments in Section~\ref{ssec:rnn}.

\textbf{Architecture Operations.} In stage (1), DARTS trained the shared weights network with the zero operation included in the list of considered operations but removed the zero operation when selecting the final architecture to evaluate in stages (2) and (3).  For our random search with weight-sharing, we decided to exclude the zero operation for both search and evaluation.

\textbf{Stage 3 Procedure.} For stage (3) evaluation, we follow the ArXiv version of DARTS \citep{dartsarxiv}, which reported two sets of results, one after training for 1600 epochs and another fine tuned result after training for an additional 1000 epochs.  In the ICLR version, \citet{liu2018darts} simply say they trained the final network to convergence.  We trained for another 1000 epochs for a total of 3600 epochs to approximate training to convergence.
\subsection{CIFAR-10 Benchmark}
\label{ssec:appendix_reproduce}
In this section, we provide additional detail for the experiments in Section~\ref{ssec:cnn}.

\textbf{Architecture Operations.} In stage (1), DARTS trained the shared weights network with the zero operation included in the list of considered operations but removed the zero operation when selecting the final architecture to evaluate in stages (2) and (3).  For our random search with weight-sharing, we decided to \emph{include} the zero operation for both search and evaluation.

\textbf{Stage 1 Procedure.}  For random search with weight-sharing, after the shared weights are fully trained, we evaluate randomly sampled architectures using the shared weights and select the best one for stage (2) evaluation.  Due to the higher cost of evaluating on the full validation set, we evaluate each architecture using 10 minibatches instead.  We split the total number of architectures to be evaluated into sets of 1000.  For each 1000, we select the best 10 according the cheap evaluation on part of the validation set and evaluate on the full validation set.  Then we select the top architecture across all sets of 1000 for stage (2) evaluation.

\textbf{Reproducibility.}
The code released by \citet{liu2018darts} did not produce deterministic results for the CNN benchmark due to non-determinism in CuDNN and in data loading.  We removed the non-deterministic behavior in CuDNN by setting
\begin{minted}{python}
cudnn.benchmark = False
cudnn.deterministic = True
cudnn.enabled=True
\end{minted}
Note that this only disables the non-deterministic functions in CuDNN and does not adversely affect training time as much as turning off CuDNN completely.  We fix additional non-determinism from data loading by setting the seed for the \texttt{random} package in addition to \texttt{numpy.random} and \texttt{pytorch} seed and turning off multiple threads for data loading.

We ran ASHA and one set of trials for Random Search (5) with weight-sharing using the non-deterministic code before fixing the seeding to get deterministic results; all other settings for random search with weight-sharing are deterministic.  Hence, the result for ASHA does not satisfy exact reproduciblity due to non-deterministic training and asynchronous updates.

\bibliographystyle{abbrvnat}
\bibliography{nas}

\end{document}